\documentclass[11pt]{article}
\usepackage{preprint}
\usepackage{colortbl,etoc,longtable,needspace,placeins,tikz,wrapfig}
\usetikzlibrary{arrows.meta,positioning,decorations.pathreplacing,shapes.geometric}

\definecolor{resultblue}{HTML}{376B8D}
\definecolor{resultorange}{HTML}{C27B38}
\definecolor{resultteal}{HTML}{588E7B}
\definecolor{resultgray}{HTML}{9EA7AD}
\definecolor{resultslate}{HTML}{3E4C59}
\colorlet{resultshade}{resultblue!8}
\colorlet{resultheader}{black!5}

\preprinttitle{EgoUseBench}{Benchmarking Computer-Use Agents under Evolving First-Person Contexts}
\author{Yurun Chen, Biao Yi, Qihang Yu, Keting Yin, Shengyu Zhang}
\renewcommand{\PreprintAffiliation}{Zhejiang University}
\date{}
\preprintlogos
  {\preprintlogo[width=.27\linewidth]{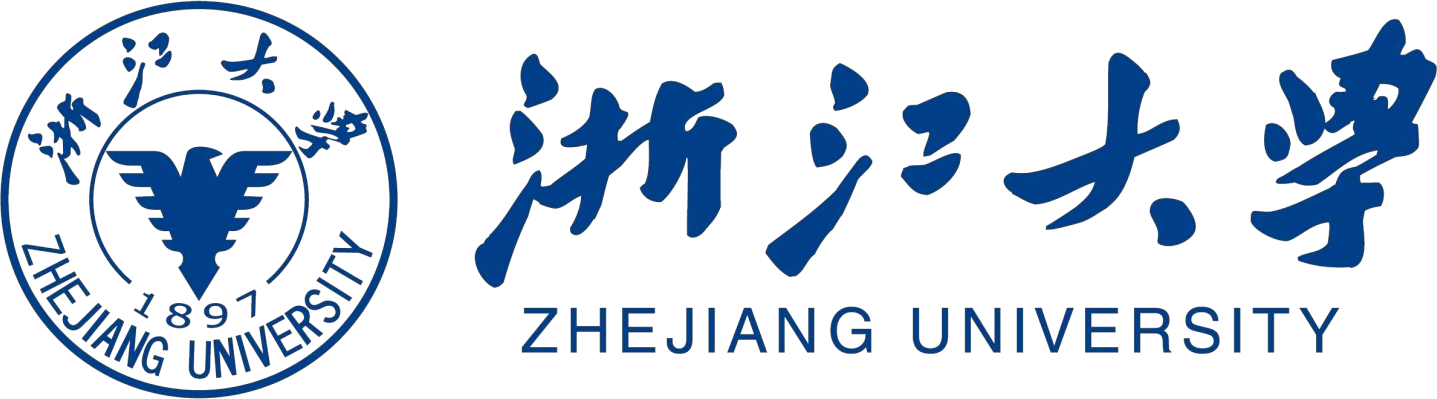}}
  {\preprintlogo[width=.49\linewidth]{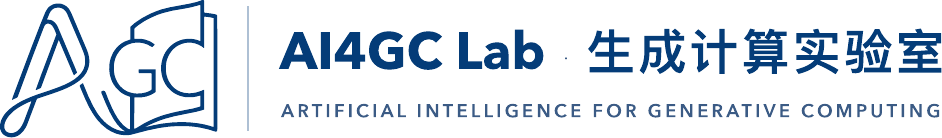}}
\renewcommand{\PreprintProjectMark}{%
  \preprintlogo[height=48pt,trim=77bp 46bp 109bp 44bp,clip]{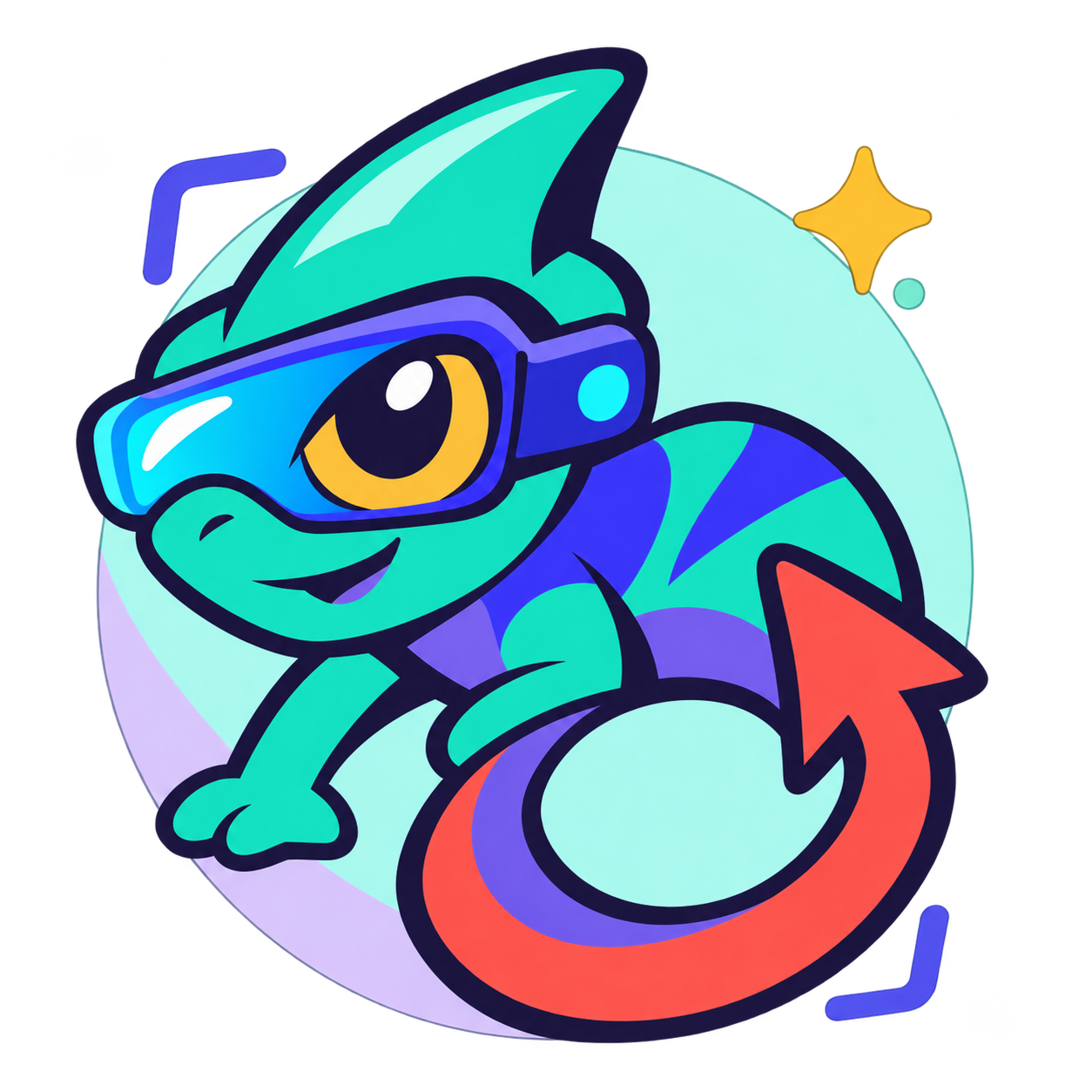}}
\preprintresources
  {https://yurunchen.github.io/egousebench/}
  {https://github.com/YurunChen/EgoUseBench}
  {https://huggingface.co/datasets/yurun-chen/EgoUseBench}
\preprintpreviewtrue

\begin{document}
\etocdepthtag.toc{main}
\begin{preprintfrontmatter}
  \maketitle
  \begin{abstract}
AI wearables, such as smart glasses, extend the context available to agents beyond digital environments to users' everyday physical surroundings. With these devices, computer tasks can originate from users' ongoing real-world activities rather than solely from explicit instructions. Through first-person observations, agents can identify objects and activities relevant to user requests and ground digital tasks in the surrounding context. We call this interaction paradigm \textbf{Ego Use}. However, first-person observations are limited by the wearer's viewpoint and may provide incomplete or ambiguous evidence of ongoing activities. As new observations arrive, agents may need to revise their understanding of the physical context and user intent, and reassess whether prior digital actions remain valid. Whether current agents can reliably handle these changes remains insufficiently evaluated.

We introduce \textsc{EgoUseBench}, a benchmark for evaluating Ego Use across three interaction settings with different levels of first-person context. Computer-Use relies on explicit instructions, Offline Ego Use provides prerecorded first-person context before execution, and Online Ego Use provides continuous observations as user intent evolves. Based on over 74 hours of real-world first-person recordings, we construct 693 tasks across the three interaction settings. To capture task evolution, we track first-person observations, user requirements, and website state separately. We further group tasks by how information changes during execution, resulting in three types: Graft (revised user requirements), Refresh (new first-person observations), and Clarify (agent-initiated information seeking). We evaluate several advanced models and find that average success rates in Online Ego Use are 26.3\% lower than in Computer-Use. This result reveals the difficulty of keeping task execution aligned with evolving first-person context and user intent.

\end{abstract}

\end{preprintfrontmatter}
\suppressfloats[t]
\section{Introduction}
\label{sec:intro}
When a user asks a personal assistant to ``\textit{find one like this,}'' the request may depend as much on what the user sees as on what they say. First-person AI wearables, such as Ray-Ban Meta and Looki L1~\cite{meta2025multimodalglasses,looki2026l1}, provide visual context to help assistants understand such requests. With this context available to the agent, users can refer directly to objects and activities around them, without describing the scene in detail or translating their needs into explicit digital instructions. The agent then grounds the user's instruction in the first-person context and uses it to guide digital actions. We define this interaction paradigm as \textit{\textbf{Ego Use}}.

\begin{figure}[tbp]
\centering
\includegraphics[width=\textwidth]{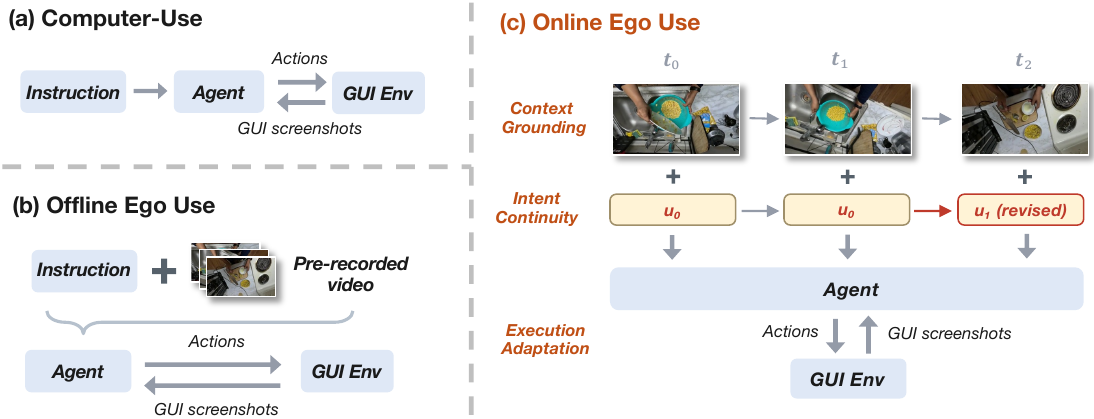}
\caption{Comparison of task-context availability. (a) Computer-Use receives an explicit instruction. (b) Offline Ego Use additionally receives prerecorded first-person context before execution. (c) Online Ego Use incorporates streaming observations and evolving user intent during execution. GUI screenshots provide interaction feedback in all three settings.}
\label{fig:overview}
\end{figure}

Figure~\ref{fig:overview} illustrates three interaction settings distinguished by the availability of first-person context during task execution. \textbf{(1) Computer-Use:} agents perform computer tasks through graphical user interfaces based on explicit user instructions, which may include descriptions of task-relevant context~\cite{deng2023mind2web,zhou2023webarena,koh2024visualwebarena,he2024webvoyager,xie2024osworld,drouin2024workarena}. \textbf{(2) Offline Ego Use:} agents additionally receive prerecorded first-person videos before execution, using them to ground subsequent digital actions~\cite{yu2026ego2web,liu2026egobench}. \textbf{(3) Online Ego Use:} agents receive continuous first-person observations during task execution, while user intent may evolve through ongoing interaction. They must therefore continuously update their understanding of the physical context and user intent, and adapt their digital actions accordingly. However, existing benchmarks provide limited evidence on reliable task execution in this dynamic first-person setting.

Online Ego Use poses challenges in grounding incomplete observations, tracking evolving user intent, and adapting ongoing actions. Consider a user preparing a snack who asks an agent to add the tortillas they are using to an online grocery cart. However, the first-person view captures only part of the package, leaving the exact product uncertain. The agent must infer the intended item from incomplete visual evidence rather than rely on an unambiguous view (\textbf{Context Grounding}). As the user subsequently asks it to add a spread as well, the agent must interpret this follow-up in relation to the ongoing task and maintain a coherent understanding of the user's evolving intent (\textbf{Intent Continuity}). Crucially, subsequent observations may reveal that the initially selected tortillas do not match what the user is actually using, requiring the agent to revisit its earlier interpretation and determine whether its previous selection should be replaced (\textbf{Execution
Adaptation}). New evidence or revised requirements can change which earlier choices remain valid, while the website retains the effects of actions already taken. This motivates our central question: \textit{Can agents reliably complete computer tasks as first-person observations and user intent evolve during execution?}

We introduce \textsc{EgoUseBench}, a benchmark built from over 74 hours of real-world first-person recordings. It comprises 693 tasks across three interaction settings, including 293 Online Ego Use tasks, 200 Offline Ego Use tasks, and 200 Computer-Use tasks. The tasks span three categories: product selection, tutorial retrieval, and place bookmarking. The benchmark has two key design features. \textbf{(1) Asynchronous State Evolution.} The environment maintains three evolving states: first-person observations, user requirements, and website state. Observations are replayed at fixed intervals, user requirements change according to predefined schedules, and website states evolve through agent actions.
\textbf{(2) Dynamic Information Updates.} We introduce three mechanisms for information updates during task execution: \textit{Graft} modifies user requirements, \textit{Refresh} provides new first-person observations for revising prior decisions, and \textit{Clarify} allows agents to seek additional information. These mechanisms capture user-driven changes, environment updates, and agent-initiated information seeking.

We conduct extensive experiments on \textsc{EgoUseBench} with multiple advanced models to evaluate their performance across different interaction settings. Average success rates are 63.4\% in Online Ego Use, 64.2\% in Offline Ego Use, and 86.1\% in Computer-Use, revealing a substantial performance gap between Ego Use and Computer-Use. We further examine how agents handle the information updates that distinguish Online from Offline Ego Use. Agents struggle more when new visual evidence invalidates an earlier choice than when it first reveals a target. More request changes do not consistently reduce success. Scene-change alerts and descriptions improve success, while longer visual histories offer no consistent benefit. These findings point to the difficulty of maintaining consistent decisions as first-person context and user intent evolve. Agents need to recognize how new information affects earlier choices and revise them accordingly.

Our contributions are threefold. \textbf{(i) Ego Use.} To our knowledge, we are the first to introduce Ego Use, in which agents use first-person observations to interpret user requests and perform computer tasks. We distinguish Offline and Online Ego Use, with the latter involving evolving observations and user intent during execution. \textbf{(ii) \textsc{EgoUseBench}.} We introduce a benchmark of 693 executable tasks across three interaction settings. It models first-person observations, user requirements, and website state as separate evolving states, and supports updates to observations and requirements as well as agent-initiated clarification. \textbf{(iii) Extensive Experiments.} We evaluate multiple advanced models through cross-setting comparisons and execution-trace analysis. Our analysis reveals how information updates affect task success and the difficulty of keeping task execution aligned with evolving first-person context and user intent.

\section{Related Work}
\label{sec:related}

\paragraph{Computer-Use.}
Computer-use agents carry out user-specified computer tasks through graphical user interfaces (GUIs). Mind2Web~\cite{deng2023mind2web} introduced a large-scale dataset for web agents. WebArena~\cite{zhou2023webarena}, VisualWebArena~\cite{koh2024visualwebarena}, and WebVoyager~\cite{he2024webvoyager} study multi-step interaction and visual grounding across realistic web environments and live websites. OSWorld~\cite{xie2024osworld} extends evaluation to desktop applications and cross-application workflows. These studies advance interface understanding, task planning, and digital execution, with agents acting primarily on user-specified goals and feedback from the digital environment.

\paragraph{Offline Ego Use.}
First-person video provides context for both understanding user needs and carrying out tasks. Ego4D~\cite{grauman2022ego4d} and EgoLife~\cite{yang2025egolife} advance long-horizon activity understanding and personal memory, while WAGIBench~\cite{veerabadran2025wagibench} studies how to infer user goals from multimodal context. Ego2Web~\cite{yu2026ego2web} grounds web tasks in prerecorded egocentric video, and EgoBench~\cite{liu2026egobench} adds tool use and multi-turn user interaction. These benchmarks connect physical observations to digital actions, but provide the visual evidence in advance rather than updating it as the computer task unfolds.

\paragraph{Online Ego Use.}
Recent work studies understanding and interaction under continuous first-person observation. EgoSAT~\cite{lei2026egosat} evaluates temporal reasoning over streaming video. Vinci2 and its EgoServe benchmark~\cite{gong2026vinci2} study proactive assistance under continuous observation. ProactEgoMem~\cite{liao2026proactegomem} focuses on information capture and state tracking in streaming video, while OmniAssistBench~\cite{sun2026omniassistbench} evaluates multi-turn interaction with multimodal assistants. These studies focus on understanding evolving context, offering timely assistance, or maintaining memory. \textsc{EgoUseBench} instead evaluates whether agents can reliably complete computer tasks while first-person visual information and user intent evolve during execution.

\section{\textsc{EgoUseBench}}
\label{sec:benchmark}
\begin{figure}[tbp]
\centering
\includegraphics[width=\textwidth]{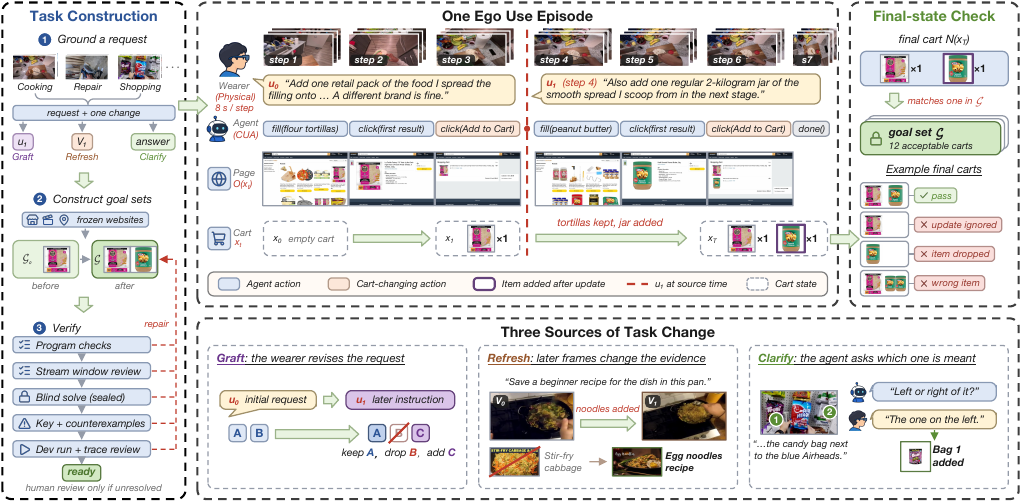}
\caption{Overview of \textsc{EgoUseBench}. The left panel shows task construction and verification, including review of the replay window. The center shows a Graft episode with 8 seconds of source footage per step. The follow-up request arrives at step 4 according to the replay timeline, independently of the prior cart action. Browser operations are shown in shorthand. The right panel compares the final cart against the acceptable goal states. The bottom panels illustrate Graft, Refresh, and Clarify.}
\label{fig:method-overview}
\end{figure}

\textsc{EgoUseBench} evaluates whether computer-use agents can complete browser tasks as first-person observations and user requests evolve. New information can change what counts as a correct outcome while the effects of earlier browser actions persist. The benchmark controls when that information arrives and lets the agent ask for clarification. We judge completion from the final website state and use the trace to examine how the agent got there (Figure~\ref{fig:method-overview}).

\subsection{Task Definition}
\label{sec:task-definition}
The prompt asks the agent to complete the wearer's request and describes the actions available across all task types: search or inspect the website, change its state, ask a question, or finish. To separate the effects of those actions from changes in the task, let $x$ denote the current website state and $\mathcal{G}$ the set of states that satisfy the resolved request. New visual evidence or a revised request can change $\mathcal{G}$ while leaving $x$ intact:
\begin{equation}
 (x,\mathcal{G}^{-})\longrightarrow(x,\mathcal{G}^{+}),
 \label{eq:goal-update}
\end{equation}
where $\mathcal{G}^{-}$ and $\mathcal{G}^{+}$ are the acceptable states before and after the update.

\textbf{Refresh} holds the request fixed. Its \emph{Replace} subtype introduces visual evidence that invalidates an earlier choice, whereas \emph{Retain} leaves that choice valid. \textbf{Graft} instead changes the user's request. \emph{Refine} adjusts the requested attributes or quantity, \emph{Augment} adds a requirement while preserving earlier ones, \emph{Replace} substitutes one requested choice for another, and \emph{Drop} withdraws a requirement.

\textbf{Clarify} presents multiple plausible referents for the wearer's intended target. A reviewed reply to an agent question resolves the ambiguity without naming a website item. \emph{Position} uses spatial location, \emph{Sequence} uses temporal order, and \emph{Variant} uses visible attributes such as color or size. Across all three task types, the agent must reach an acceptable final state within the step budget.

\subsection{Task Construction}
\label{sec:construction}

\paragraph{Design.}
We construct 293 Online Ego Use tasks, comprising 200 tasks based on Graft, Refresh, or Clarify and 93 tasks that combine them. We also construct 200 Computer-Use and 200 Offline Ego Use tasks for matched comparisons, keeping the websites and final goals fixed. Tasks are constructed from candidate activities in Ego2Web~\cite{yu2026ego2web}, CaptainCook4D~\cite{peddi2023captaincook4d}, HD-EPIC~\cite{perrett2025hdepic}, EgoOops~\cite{haneji2024egooops}, EgoLife~\cite{yang2025egolife}, HoloAssist~\cite{wang2023holoassist}, and EgoSchema~\cite{mangalam2023egoschema}. Source coverage is provided in the supplementary material. For Refresh, we select an observed change and hold the wording fixed, checking whether the later scene changes the answer. For Graft, we write a follow-up that edits a specific requirement at a plausible point in the activity. For Clarify, we select an opening scene with multiple plausible referents and write an answer that resolves the distinction. We review the entire replay window: the opening must support the request, and intervening frames must not reveal later evidence early.

The request specifies what to accomplish on the website and under what constraints. The footage supplies the referent. Before selecting listings, we decide how closely a website item must match that referent. We use only the request to search for candidates. If the results all share an attribute that the agent should infer from the footage, we broaden the search so that the catalog alone does not determine that attribute.

\paragraph{Validation.}
\noindent\textbf{(1) Acceptable outcomes.}
Let $R_0$ and $R_1$ denote the initial and final requirements, with the intended referent resolved for verification. We define
\begin{equation}
 \mathcal{G}^{(i)}=\{x\in\mathcal{X}_{\mathrm{site}}:x\models R_i\},
\end{equation}
where $\mathcal{X}_{\mathrm{site}}$ is the set of reachable states on the frozen website. An acceptable state meets the requirements without extra selections. For Graft and Refresh, we verify both sets in Eq.~\ref{eq:goal-update} to distinguish an initial error from a choice that became invalid after an update. For Clarify, we verify the goal for the resolved referent. We use public listing details to distinguish matches, mismatches, and cases with insufficient evidence. The compiler encodes verified alternatives and packaging equivalences as executable goals. Equivalent packages count toward total quantities. Minimum capacity applies to each container. Scoring uses the verified final set $\mathcal{G}=\mathcal{G}^{(1)}$.

\noindent\textbf{(2) Information shortcuts.}
After verifying the answer set, we check whether the task can be solved without the footage or without an update intended to change the goal. Let $\mathcal{H}$ be the plausible interpretations under the remaining inputs and $\mathcal{G}_h$ the acceptable states for interpretation $h$. A shortcut exists if the remaining inputs identify an outcome in $\bigcap_{h\in\mathcal{H}}\mathcal{G}_h$. A lucky guess does not establish one. If a valid answer creates a shortcut, we keep it and revise the request or choose another source.

\noindent\textbf{(3) Independent review.}
An independent Coding Agent records a solution before seeing the proposed answers, then checks answer coverage and possible shortcuts. Offline checks and a development run verify scoring, information delivery, and website execution regardless of model success. Unresolved semantic questions go to human review. Qwen3.5-Flash supports development and shortcut screening, separately from the reported models. Detailed checks are provided in the supplementary material.

\subsection{Interactive Environment}
\label{sec:environment}
\paragraph{Websites.}
We build local simulations of Amazon, YouTube, and Google Maps that reproduce the interfaces and interactions used in our tasks. The sites use publicly available product, video, and place information and support operations such as search, adding products to a cart, and saving videos or places. Agents are instructed to stay within these sites. Search ranks public fields with BM25. The rendered pages and agent-visible details use the same website data. Source annotations, answer roles, and evaluation labels are not exposed to the agent. Each episode starts from an isolated website state that persists throughout the interaction.

\paragraph{Input.}
At each decision step, the agent receives three inputs: (1) first-person observations of the wearer's surroundings, (2) the current webpage screenshot and accessibility tree, reflecting previous actions, and (3) user messages in their original wording, labeled by delivery step (Figure~\ref{fig:input-design}). Each decision permits one JSON action. Element-based actions use IDs from the current accessibility tree. Each active step advances the recording by 8\,s of source footage, independently of model latency. {Frames are sampled at 1\,fps and normally tiled four per image. Task-specific single-frame layouts and image-size limits are fixed across models and conditions.} Episodes allow up to 20 decision steps in the main evaluation and 30 in the composite evaluation (Section~\ref{sec:composite}).

\begingroup
\setlength{\intextsep}{3pt}
\begin{wrapfigure}{r}{.52\textwidth}
\centering
\includegraphics[width=\linewidth]{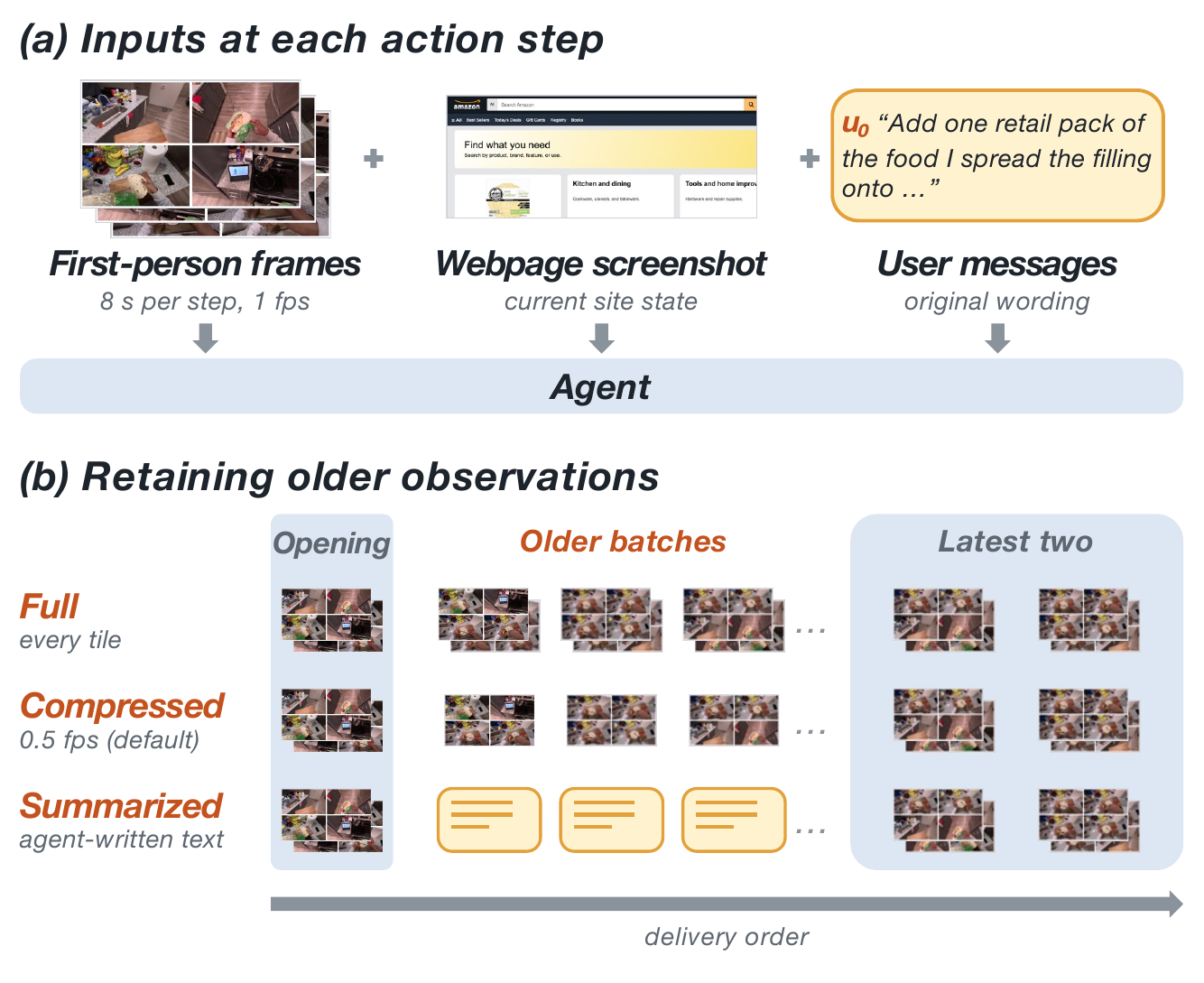}
\caption{(a) Inputs at an action step. (b) The three visual history conditions. The opening batch and the two most recent delivered batches are kept intact in all three conditions. Only the representation of older batches changes.}
\label{fig:input-design}
\end{wrapfigure}

Because information relevant to later decisions may appear in earlier observations, we retain a representation of each delivered batch, following prior work on visual-history compression in streaming assistants~\cite{song2024moviechat,wang2025streambridge}. All conditions preserve the opening batch and the two most recent batches in full. For older batches, as illustrated in Figure~\ref{fig:input-design}, we compare three history policies. (1) Full history retains all frames. (2) Compressed history, our default, {retains every other frame (0.5\,fps) using the same task-specific layout.} (3) Summarized history replaces each older batch with a short textual summary generated by the acting model alongside its action. The summary becomes available only after the corresponding batch leaves the recent-history window. None of these policies uses task annotations to select frames. The Retention comparison in Section~\ref{sec:variables} varies this policy while holding frame sampling and observation delivery fixed.
\par
\endgroup

\paragraph{Delivery.}
The replay schedule sets when new information becomes available. Whether the agent is still working determines how it receives that information. \textbf{(1) Concurrent change.} During active execution, a Graft follow-up arrives with the batch containing its reviewed onset, and a Refresh change arrives with the later scene. The follow-up in Figure~\ref{fig:method-overview} arrives at step 4. To give the agent time to act after a Refresh change, we retain only the final 16\,s of longer waits between scenes and exclude windows whose opening scene alone makes the change too late. The recording does not loop. \textbf{(2) Wake-up change.} If the agent finishes before a scheduled Graft follow-up, the episode enters standby. The follow-up resumes execution with the same website state. Standby consumes no action steps, and skipped batches are logged without being delivered. Later Refresh frames do not resume a finished episode.

\paragraph{Questions.}
All Online task types allow a question. The first question in Clarify returns the reviewed reply before the next decision. In Graft and Refresh it returns no additional task information. Questions consume decision steps. The reply is fixed rather than generated from the question, so this interface measures whether the agent seeks clarification, not the quality of its wording. The supplementary material lists the full action interface and fixed replies.

\subsection{Task Evaluation}
\label{sec:task-evaluation}
We score the final website state against the final task requirements. In Figure~\ref{fig:method-overview}, the cart must contain the requested tortilla and spread in the correct quantities, with no extras. An agent can wait for the follow-up before selecting either item or act on the initial request and then update the cart. Both paths earn the same score if they reach an acceptable state.

Let $x_T$ be the website state at termination, including when the agent has resumed from standby. We compare its normalized representation $N(x_T)$ with each verified goal. For any state $x$, $N(x)$ contains the selected entities, their quantities, and the required fields. Normalization ignores display order and converts registered package equivalents to the requested units. Success is
\begin{equation}
 \operatorname{Success}(x_T)=
 \mathbf{1}\left[\exists g\in\mathcal{G}:N(x_T)=N(g)\right].
 \label{eq:success}
\end{equation}
Intermediate mistakes do not count against an agent that corrects them by termination. Success does not require a clarification question or an extra submit action. We compute success rates over valid runs and report infrastructure failures and runs without terminal evidence separately.

To analyze responses to an update, we compare its actual delivery step with the first selection and inspect the website state on either side of the update. This separates first selections from revisions and retained choices. In a Refresh Retain task, a valid final choice earns credit, but the score alone does not show whether the agent used the later frames.

\section{Experiments}
\label{sec:experiments}
Our evaluation addresses four research questions: \textbf{RQ1:} How reliably do agents complete Online Ego Use tasks? \textbf{RQ2:} How do success rates and token costs differ across Computer-Use, Offline Ego Use, and Online Ego Use? \textbf{RQ3:} Which online input conditions affect task completion? \textbf{RQ4:} How do agents perform under combinations of information updates?

\subsection{Experimental Setup}
\label{sec:setup}
\mbox{\textbf{(1) Models.}} {We evaluate GPT-6.1 Terra~\cite{openai2026terra}, Gemini 3.8 Flash~\cite{deepmind2026gemini38flash}, Doubao Seed 2.1 Pro~\cite{bytedance2026seed21}, DeepSeek-V4.1-Flash~\cite{deepseek2026v41flash,deepseek2026flash1017}, Kimi K2.6~\cite{moonshot2026kimik26card}, and qwen3.5-27b~\cite{qwen2026qwen35card} using a shared BrowserGym action interface. We also evaluate the smaller UI-TARS-1.5-7B, with results in the supplementary material.} \mbox{\textbf{(2) Evaluation protocol.}} The main Online evaluation uses 200 tasks from distinct recordings: 142 Graft, 30 Refresh, and 28 Clarify. Each task is run once per condition with up to 20 decision steps. Replay and compressed history follow Section~\ref{sec:environment}. \mbox{\textbf{(3) Metrics.}} Our primary metric is success rate (SR), the fraction of valid runs whose final website state satisfies Eq.~\ref{eq:success}. Overall SR weights tasks equally. Paired comparisons report SR differences ($\Delta$SR, \%) with 95\% confidence intervals.

\paragraph{Task quality.}
To validate task quality, two independent reviewers examined 40 stratified tasks, including 32 main and 8 composite tasks. They checked whether the recordings supported the requests and whether the answer sets were correct and covered reasonable alternatives. {Tasks accepted by both reviewers account for 90\% of the sample.} Detailed review criteria and task-type counts are provided in the supplementary material.

\subsection{RQ1: Online Task Completion}
\label{sec:main-results}
{We evaluate six models on the same 200 Online Ego Use tasks in Table~\ref{tab:main}. Overall SR ranges from 39.0\% to 87.0\%, with Terra leading across all three task types. Most models score higher on Graft than Refresh. On Clarify, qwen3.5-27b outperforms DeepSeek despite lower overall SR. Its 41.0\% overall SR is also close to Kimi K2.6's 39.0\% despite a much smaller total parameter count~\cite{qwen2026qwen35card,moonshot2026kimik26card}.}

\begin{table}[!htbp]

\centering
\caption{Online success rates (\%) by task type. The Tasks row gives counts shared by all models. Variant groups visible-attribute references. Overall weights scored tasks equally.}
\label{tab:main}
\small
\renewcommand{\arraystretch}{1.1}
\setlength{\tabcolsep}{1.5pt}
\begin{tabular*}{\textwidth}{@{\extracolsep{\fill}}l*{9}{r}>{\columncolor{resultshade}}r@{}}
\toprule
& \multicolumn{4}{c}{\textbf{Graft}} & \multicolumn{2}{c}{\textbf{Refresh}} & \multicolumn{3}{c}{\textbf{Clarify}} & \cellcolor{white}\\
\cmidrule(lr){2-5}\cmidrule(lr){6-7}\cmidrule(lr){8-10}
\textbf{Model} & Refine & Augment & Replace & Drop & Replace & Retain & Position & Sequence & Variant & \textbf{Overall}\\
Tasks & 73 & 47 & 14 & 8 & 15 & 15 & 12 & 11 & 5 & 200\\
\midrule
\rowcolor{resultheader}\multicolumn{11}{@{}l@{}}{\textbf{Closed-weight models}}\\
GPT-6.1 Terra~\cite{openai2026terra} & 86.3 & 89.4 & 92.9 & 100.0 & 86.7 & 80.0 & 75.0 & 90.9 & 80.0 & 87.0\\
Gemini 3.8 Flash~\cite{deepmind2026gemini38flash} & 72.6 & 78.7 & 78.6 & 87.5 & 73.3 & 80.0 & 75.0 & 81.8 & 60.0 & 76.0\\
Doubao Seed 2.1 Pro~\cite{bytedance2026seed21} & 76.7 & 76.6 & 85.7 & 87.5 & 53.3 & 73.3 & 66.7 & 72.7 & 80.0 & 75.0\\
\addlinespace
\rowcolor{resultheader}\multicolumn{11}{@{}l@{}}{\textbf{Open-weight models}}\\
DeepSeek-V4.1-Flash~\cite{deepseek2026v41flash} & 69.9 & 59.6 & 85.7 & 87.5 & 53.3 & 33.3 & 41.7 & 54.5 & 60.0 & 62.5\\
Kimi K2.6~\cite{moonshot2026kimik26card} & 39.7 & 38.3 & 57.1 & 50.0 & 6.7 & 46.7 & 16.7 & 63.6 & 40.0 & 39.0\\
qwen3.5-27b~\cite{qwen2026qwen35card} & 37.0 & 34.0 & 50.0 & 62.5 & 26.7 & 46.7 & 58.3 & 54.5 & 60.0 & 41.0\\
\bottomrule
\end{tabular*}
\end{table}

\subsection{RQ2: Interaction Settings}
\label{sec:ablation}
{Figure~\ref{fig:settings} compares the three interaction settings on the same 200 tasks for Terra, Gemini, Doubao, DeepSeek, Kimi, and qwen3.5-27b. Cross-model means weight these six models equally. Computer-Use provides the resolved final requirements as text and disables questions. Offline Ego Use provides the complete recording and original requests before execution, whereas Online Ego Use delivers them during execution. The website, starting state, browser actions, step limit, and final goal remain fixed. Clarify replies are available in both Ego Use settings.}

\begin{figure}[!htbp]
\centering\begin{minipage}[t]{.53\linewidth}\vspace{0pt}\centering
\resizebox{.98\linewidth}{!}{%
\begin{tikzpicture}[x=1cm,y=.052cm,font=\footnotesize,
  trend/.style={line width=1pt,line join=round,line cap=round},
  val/.style={font=\footnotesize,inner sep=1.5pt},
  lead/.style={line width=.4pt},
  mgemini/.style={circle,fill=resultblue,draw=white,line width=.5pt,inner sep=0pt,minimum size=4.8pt},
  mdeep/.style={rectangle,fill=resultorange,draw=white,line width=.5pt,inner sep=0pt,minimum size=4.4pt},
  mkimi/.style={regular polygon,regular polygon sides=3,fill=white,draw=resultteal,line width=.8pt,inner sep=0pt,minimum size=6pt},
  mqwen/.style={diamond,fill=white,draw=black!60,line width=.8pt,inner sep=0pt,minimum size=5.4pt}]
  \useasboundingbox (-.5,15.5) rectangle (8.55,118.5);
  \node[anchor=west,font=\small\bfseries] at (-.5,114) {(a) Success across settings};
  \fill[resultshade] (4.34,32) rectangle (5.70,109);
  \node[font=\footnotesize\itshape,text=resultblue] at (5.02,105.2) {main setting};
  \foreach \y in {40,60,80,100} {
    \draw[black!10,line width=.4pt] (0,\y) -- (5.70,\y);
    \node[anchor=east,text=black!55] at (-.06,\y) {\y};
  }
  \draw[black!40,line width=.4pt] (0,32) -- (5.70,32);
  \node[anchor=north] at (.92,31) {Computer-Use};
  \node[anchor=north] at (2.97,31) {Offline};
  \node[anchor=north] at (5.02,31) {Online};
  \draw[decorate,decoration={brace,mirror,amplitude=2.5pt},black!45,line width=.4pt] (2.42,24.5) -- (5.57,24.5);
  \node[anchor=north,text=black!60,font=\footnotesize\itshape] at (3.995,21.5) {Ego Use};
  \draw[resultslate,trend] (.92,94) -- (2.97,84) -- (5.02,87);
  \node[circle,fill=white,draw=resultslate,line width=.8pt,inner sep=0pt,minimum size=4.8pt] at (0.92,94) {};
  \node[circle,fill=white,draw=resultslate,line width=.8pt,inner sep=0pt,minimum size=4.8pt] at (2.97,84) {};
  \node[circle,fill=white,draw=resultslate,line width=.8pt,inner sep=0pt,minimum size=4.8pt] at (5.02,87) {};
  \draw[lead,resultslate] (5.10,87) -- (5.78,92);
  \node[anchor=west,text=resultslate,inner sep=1pt] at (5.80,92) {\textbf{87.0}\,Terra};
  \draw[resultblue,trend] (.92,89.5) -- (2.97,76.5) -- (5.02,76);
  \node[circle,fill=white,draw=resultblue,line width=.8pt,inner sep=0pt,minimum size=4.8pt] at (0.92,89.5) {};
  \node[circle,fill=white,draw=resultblue,line width=.8pt,inner sep=0pt,minimum size=4.8pt] at (2.97,76.5) {};
  \node[circle,fill=white,draw=resultblue,line width=.8pt,inner sep=0pt,minimum size=4.8pt] at (5.02,76) {};
  \draw[lead,resultblue] (5.10,76) -- (5.78,81);
  \node[anchor=west,text=resultblue,inner sep=1pt] at (5.80,81) {\textbf{76.0}\,Gemini};
  \draw[violet!70!black,trend] (.92,92) -- (2.97,76.5) -- (5.02,75);
  \node[regular polygon,regular polygon sides=5,fill=white,draw=violet!70!black,line width=.8pt,inner sep=0pt,minimum size=4.8pt] at (0.92,92) {};
  \node[regular polygon,regular polygon sides=5,fill=white,draw=violet!70!black,line width=.8pt,inner sep=0pt,minimum size=4.8pt] at (2.97,76.5) {};
  \node[regular polygon,regular polygon sides=5,fill=white,draw=violet!70!black,line width=.8pt,inner sep=0pt,minimum size=4.8pt] at (5.02,75) {};
  \draw[lead,violet!70!black] (5.10,75) -- (5.78,72);
  \node[anchor=west,text=violet!70!black,inner sep=1pt] at (5.80,72) {\textbf{75.0}\,Doubao};
  \draw[resultorange,trend] (.92,89) -- (2.97,66) -- (5.02,62.5);
  \node[rectangle,fill=white,draw=resultorange,line width=.8pt,inner sep=0pt,minimum size=4.8pt] at (0.92,89) {};
  \node[rectangle,fill=white,draw=resultorange,line width=.8pt,inner sep=0pt,minimum size=4.8pt] at (2.97,66) {};
  \node[rectangle,fill=white,draw=resultorange,line width=.8pt,inner sep=0pt,minimum size=4.8pt] at (5.02,62.5) {};
  \draw[lead,resultorange] (5.10,62.5) -- (5.78,62.5);
  \node[anchor=west,text=resultorange,inner sep=1pt] at (5.80,62.5) {\textbf{62.5}\,DeepSeek};
  \draw[resultteal,trend,dashed] (.92,77.5) -- (2.97,40.5) -- (5.02,39);
  \node[regular polygon,regular polygon sides=3,fill=white,draw=resultteal,line width=.8pt,inner sep=0pt,minimum size=4.8pt] at (0.92,77.5) {};
  \node[regular polygon,regular polygon sides=3,fill=white,draw=resultteal,line width=.8pt,inner sep=0pt,minimum size=4.8pt] at (2.97,40.5) {};
  \node[regular polygon,regular polygon sides=3,fill=white,draw=resultteal,line width=.8pt,inner sep=0pt,minimum size=4.8pt] at (5.02,39) {};
  \draw[lead,resultteal] (5.10,39) -- (5.78,35.8);
  \node[anchor=west,text=resultteal,inner sep=1pt] at (5.80,35.8) {\textbf{39.0}\,Kimi};
  \draw[black!60,trend,dotted] (.92,74.5) -- (2.97,41.5) -- (5.02,41);
  \node[diamond,fill=white,draw=black!60,line width=.8pt,inner sep=0pt,minimum size=4.8pt] at (0.92,74.5) {};
  \node[diamond,fill=white,draw=black!60,line width=.8pt,inner sep=0pt,minimum size=4.8pt] at (2.97,41.5) {};
  \node[diamond,fill=white,draw=black!60,line width=.8pt,inner sep=0pt,minimum size=4.8pt] at (5.02,41) {};
  \draw[lead,black!60] (5.10,41) -- (5.78,44.2);
  \node[anchor=west,text=black!60,inner sep=1pt] at (5.80,44.2) {\textbf{41.0}\,qwen3.5-27b};
\end{tikzpicture}%
}%
\end{minipage}\hfill\begin{minipage}[t]{.45\linewidth}\vspace{0pt}
\raggedright\small\textbf{(b) Total tokens per task (k)}\par\smallskip
\centering\footnotesize
\renewcommand{\arraystretch}{1.52}
\setlength{\tabcolsep}{2pt}
\begin{tabularx}{\linewidth}{@{}l>{\centering\arraybackslash}X>{\centering\arraybackslash}X>{\columncolor{resultshade}\centering\arraybackslash}Xr@{}}
\toprule
Model & \makebox[\linewidth][c]{Computer-Use} & Offline & Online & $\Delta$ (\%)\\
\midrule
Terra & 126.3 & 433.0 & 298.5 & $-31.0$\\
Gemini & 179.7 & 512.2 & 317.6 & $-38.0$\\
Doubao & 105.3 & 394.3 & 250.1 & $-36.6$\\
DeepSeek & 178.2 & 347.3 & 290.2 & $-16.4$\\
Kimi & 84.2 & 328.3 & 226.8 & $-30.9$\\
qwen3.5-27b & 132.6 & 425.5 & 288.6 & $-32.2$\\
\bottomrule
\end{tabularx}
\end{minipage}
\caption{Interaction settings on the same 200 tasks per model. (a) SR (\%). (b) Mean total tokens per task across successful and failed runs, in thousands. $\Delta$ gives the relative token change (\%) from Offline to Online. Shading marks Online Ego Use.}
\label{fig:settings}

\end{figure}

\newcommand{\MatchedOnlineColor}{}
\newcommand{\MatchedOfflineLabel}{Offline Ego Use}
\newcommand{\MatchedOnlineLabel}{Online Ego Use}
\newcommand{\MatchedSRTerra}{%
  \MatchedSR{Terra}{Computer-Use}{92.3}{89.0 & 95.7 & 92.9 & 100.0}{93.3}{100.0}{100.0}{100.0 & 100.0 & 100.0}{94.0}%
  \MatchedSR{}{\MatchedOfflineLabel}{88.7}{89.0 & 89.4 & 78.6 & 100.0}{86.7}{60.0}{71.4}{66.7 & 81.8 & 60.0}{84.0}%
  \MatchedOnlineColor\MatchedSR{}{\MatchedOnlineLabel}{88.7}{86.3 & 89.4 & 92.9 & 100.0}{86.7}{80.0}{82.1}{75.0 & 90.9 & 80.0}{87.0}%
}
\newcommand{\MatchedSRGemini}{%
  \MatchedSR{Gemini}{Computer-Use}{86.6}{83.6 & 85.1 & 100.0 & 100.0}{93.3}{93.3}{100.0}{100.0 & 100.0 & 100.0}{89.5}%
  \MatchedSR{}{\MatchedOfflineLabel}{78.2}{79.5 & 74.5 & 85.7 & 75.0}{80.0}{66.7}{71.4}{58.3 & 81.8 & 80.0}{76.5}%
  \MatchedOnlineColor\MatchedSR{}{\MatchedOnlineLabel}{76.1}{72.6 & 78.7 & 78.6 & 87.5}{73.3}{80.0}{75.0}{75.0 & 81.8 & 60.0}{76.0}%
}
\newcommand{\MatchedSRDoubao}{%
  \MatchedSR{Doubao}{Computer-Use}{90.8}{87.7 & 93.6 & 92.9 & 100.0}{93.3}{93.3}{96.4}{91.7 & 100.0 & 100.0}{92.0}%
  \MatchedSR{}{\MatchedOfflineLabel}{81.7}{83.6 & 78.7 & 85.7 & 75.0}{86.7}{66.7}{50.0}{25.0 & 63.6 & 80.0}{76.5}%
  \MatchedOnlineColor\MatchedSR{}{\MatchedOnlineLabel}{78.2}{76.7 & 76.6 & 85.7 & 87.5}{53.3}{73.3}{71.4}{66.7 & 72.7 & 80.0}{75.0}%
}
\newcommand{\MatchedSRDeepSeek}{%
  \MatchedSR{DeepSeek}{Computer-Use}{86.6}{83.6 & 87.2 & 92.9 & 100.0}{86.7}{93.3}{100.0}{100.0 & 100.0 & 100.0}{89.0}%
  \MatchedSR{}{\MatchedOfflineLabel}{68.3}{71.2 & 61.7 & 64.3 & 87.5}{60.0}{46.7}{67.9}{66.7 & 72.7 & 60.0}{66.0}%
  \MatchedOnlineColor\MatchedSR{}{\MatchedOnlineLabel}{69.0}{69.9 & 59.6 & 85.7 & 87.5}{53.3}{33.3}{50.0}{41.7 & 54.5 & 60.0}{62.5}%
}
\newcommand{\MatchedSRKimi}{%
  \MatchedSR{Kimi}{Computer-Use}{75.4}{72.6 & 74.5 & 85.7 & 87.5}{73.3}{73.3}{92.9}{83.3 & 100.0 & 100.0}{77.5}%
  \MatchedSR{}{\MatchedOfflineLabel}{43.0}{41.1 & 44.7 & 57.1 & 25.0}{26.7}{33.3}{39.3}{16.7 & 72.7 & 20.0}{40.5}%
  \MatchedOnlineColor\MatchedSR{}{\MatchedOnlineLabel}{41.5}{39.7 & 38.3 & 57.1 & 50.0}{6.7}{46.7}{39.3}{16.7 & 63.6 & 40.0}{39.0}%
}
\newcommand{\MatchedSRQwen}{%
  \MatchedSR{qwen3.5-27b}{Computer-Use}{70.4}{69.9 & 70.2 & 78.6 & 62.5}{73.3}{80.0}{92.9}{91.7 & 90.9 & 100.0}{74.5}%
  \MatchedSR{}{\MatchedOfflineLabel}{43.0}{39.7 & 42.6 & 35.7 & 87.5}{26.7}{46.7}{39.3}{25.0 & 54.5 & 40.0}{41.5}%
  \MatchedOnlineColor\MatchedSR{}{\MatchedOnlineLabel}{38.7}{37.0 & 34.0 & 50.0 & 62.5}{26.7}{46.7}{57.1}{58.3 & 54.5 & 60.0}{41.0}%
}

\begin{table}[!htbp]
\centering
\caption{Success rates (\%) across interaction settings on the same 200 tasks per model. Offline and Online denote the two Ego Use settings. Table~\ref{tab:main} gives subtype counts.}
\label{tab:matched-settings-sr}
\small
\renewcommand{\arraystretch}{1.10}
\setlength{\heavyrulewidth}{.70pt}
\setlength{\lightrulewidth}{.40pt}
\setlength{\cmidrulewidth}{.30pt}
\setlength{\tabcolsep}{2.2pt}
\renewcommand{\MatchedOfflineLabel}{Offline}
\renewcommand{\MatchedOnlineLabel}{Online}
\newcommand{\MatchedSR}[9]{\textbf{#1} & #2 & #4 & #5 & #6 & #8 & #9\\}
\begin{tabular*}{\linewidth}{@{\extracolsep{\fill}}ll*{9}{r}>{\columncolor{resultshade}}r@{}}
\toprule
& & \multicolumn{4}{c}{\textbf{Graft}} & \multicolumn{2}{c}{\textbf{Refresh}} & \multicolumn{3}{c}{\textbf{Clarify}} & \cellcolor{white}\\
\cmidrule(lr){3-6}\cmidrule(lr){7-8}\cmidrule(lr){9-11}
\textbf{Model} & \textbf{Setting} & Refine & Augment & Replace & Drop & Replace & Retain & Position & Sequence & Variant & \textbf{Overall}\\
\midrule
\MatchedSRTerra
\addlinespace[2.5pt]
\MatchedSRGemini
\addlinespace[2.5pt]
\MatchedSRDoubao
\midrule
\MatchedSRDeepSeek
\addlinespace[2.5pt]
\MatchedSRKimi
\addlinespace[2.5pt]
\MatchedSRQwen
\bottomrule
\end{tabular*}
\end{table}

{Figure~\ref{fig:settings}a shows that both Ego Use settings trail Computer-Use for all six models, while their average success rates are close. Table~\ref{tab:matched-settings-sr} breaks down these comparisons by task subtype. Relative to Computer-Use, Graft has the smallest average drop of the three task types in both Ego Use settings. Clarify--Position shows the largest decline of any subtype in both settings. Even with the full recording, agents still have to determine which object a spatial reference denotes.}

{Moving from Offline to Online, Refresh--Replace shows the largest average decline: four models lose ground and none improve. Graft--Replace instead improves on average. The contrast highlights what visually driven revision adds: agents must recognize from new observations that an earlier choice no longer fits, without a new request. Clarify--Position shows the largest average gain, helping to offset losses elsewhere. Table~\ref{tab:setting-pairs} reports paired outcomes by task type.}

\paragraph{{Token Cost.}}
{Figure~\ref{fig:settings}b reports mean total tokens per task across successful and failed runs. Online uses 16.4\% to 38.0\% fewer tokens than Offline, consistent with gradual observation delivery and compressed visual history. The supplementary material separates input and output tokens and reports costs on tasks successful in both settings.}

\subsection{RQ3: Online Input Conditions}
\label{sec:variables}
{Table~\ref{tab:variables} compares Terra, DeepSeek, and Kimi under three online input conditions: update delivery mode (Arrival), visual history representation (Retention), and scene-change information (Cue). Each comparison uses the same tasks across conditions. The supplementary material further compares replay pace by varying the source-video duration advanced per decision step.}

\begin{table}[!htbp]

\centering
\caption{{Success rates (\%) under online input conditions. Arrival uses Graft tasks and Cue uses Refresh tasks. Each comparison uses the same tasks across conditions. Shading marks the default condition.}}
\label{tab:variables}
\small
\renewcommand{\arraystretch}{1.1}
\setlength{\tabcolsep}{2pt}
\begin{minipage}[t]{.275\linewidth}
\textbf{(a) Arrival}\hfill 98 tasks\par\smallskip
\begin{tabularx}{\linewidth}{@{}>{\raggedright\arraybackslash}X>{\columncolor{resultshade}}rr@{}}
\toprule
Model & Concurrent & Wake-up\\
\midrule
Terra & 89.8 & 81.6\\
DeepSeek & 73.5 & 53.1\\
Kimi & 44.9 & 29.6\\
\bottomrule
\end{tabularx}
\end{minipage}\hfill
\begin{minipage}[t]{.369\linewidth}
\textbf{(b) Retention}\hfill 47 tasks\par\smallskip
\begin{tabularx}{\linewidth}{@{}>{\raggedright\arraybackslash}Xr>{\columncolor{resultshade}}rr@{}}
\toprule
Model & Full & Compressed & Summarized\\
\midrule
Terra & 85.1 & 89.4 & 85.1\\
DeepSeek & 57.4 & 63.8 & 48.9\\
Kimi & 42.6 & 36.2 & 48.9\\
\bottomrule
\end{tabularx}
\end{minipage}\hfill
\begin{minipage}[t]{.326\linewidth}
\textbf{(c) Cue}\hfill 30 tasks\par\smallskip
\begin{tabularx}{\linewidth}{@{}>{\raggedright\arraybackslash}X>{\columncolor{resultshade}}rrr@{}}
\toprule
Model & Silent & Cued & Cued + facts\\
\midrule
Terra & 83.3 & 86.7 & 86.7\\
DeepSeek & 43.3 & 50.0 & 56.7\\
Kimi & 26.7 & 26.7 & 30.0\\
\bottomrule
\end{tabularx}
\end{minipage}
\end{table}

\paragraph{Arrival.}
{Table~\ref{tab:variables}a compares Concurrent and Wake-up changes on the same Graft tasks. Both retain website state and share a 20-step budget. Concurrent has higher SR across all three models. The largest gap occurs for DeepSeek, which scores 73.5\% versus 53.1\%. In most tasks where DeepSeek succeeds with Concurrent but fails with Wake-up, it finds a valid new answer but retains an outdated selection.}

\paragraph{Retention.}
{Table~\ref{tab:variables}b compares Full, Compressed, and Summarized visual histories under the same replay. More visual history does not consistently improve success. Compressed scores highest for Terra and DeepSeek, while Summarized scores highest for Kimi. Supplementary traces show how missing or inaccurate summaries hinder later corrections once the original frames are removed.}

\paragraph{Cue.}
{Table~\ref{tab:variables}c compares Silent, Cued, and Cued + facts with frames and delivery times fixed. Cued adds a scene-change alert to Silent, and Cued + facts also supplies verified scene descriptions. Alerts modestly improve or match SR. Descriptions bring further overall gains for DeepSeek and Kimi, but can also introduce errors. Relative to Cued, DeepSeek improves on Replace but declines on Retain, sometimes selecting a newly visible object instead of the requested one.}

\Needspace{13\baselineskip}
\subsection{RQ4: Composite Changes}
\label{sec:composite}
\begingroup
\setlength{\intextsep}{3pt}
\begin{wraptable}{r}{.52\textwidth}

\centering
\caption{{Online success rates (\%) on the 93 composite tasks.}}
\label{tab:composite}
\small
\renewcommand{\arraystretch}{1.1}
\setlength{\tabcolsep}{2pt}
\begin{tabularx}{\linewidth}{@{}Xrrr@{}}
\toprule
\textbf{Combination} & Tasks & \textbf{Gemini} & \textbf{DeepSeek}\\
\midrule
Refresh + Graft & 25 & 76.0 & 56.0\\
Clarify + Graft & 25 & 92.0 & 44.0\\
Clarify + Refresh & 21 & 61.9 & 23.8\\
Clarify + Refresh + Graft & 22 & 77.3 & 36.4\\
\midrule
\rowcolor{resultshade}Overall & 93 & 77.4 & 40.9\\
\bottomrule
\end{tabularx}
\end{wraptable}

Table~\ref{tab:composite} compares Gemini and DeepSeek on 93 tasks combining Graft, Refresh, and Clarify in four ways. Both models use the same tasks, websites, observation windows, update times, and 30-step budgets. Gemini achieves 77.4\% overall SR and DeepSeek achieves 40.9\%. Supplementary traces show agents responding to new information while leaving earlier decisions unresolved. They may select newly revealed items without resolving the initial reference, or search for a revised target while retaining an outdated selection.
\par
\endgroup

\begingroup
\setlength{\intextsep}{3pt}
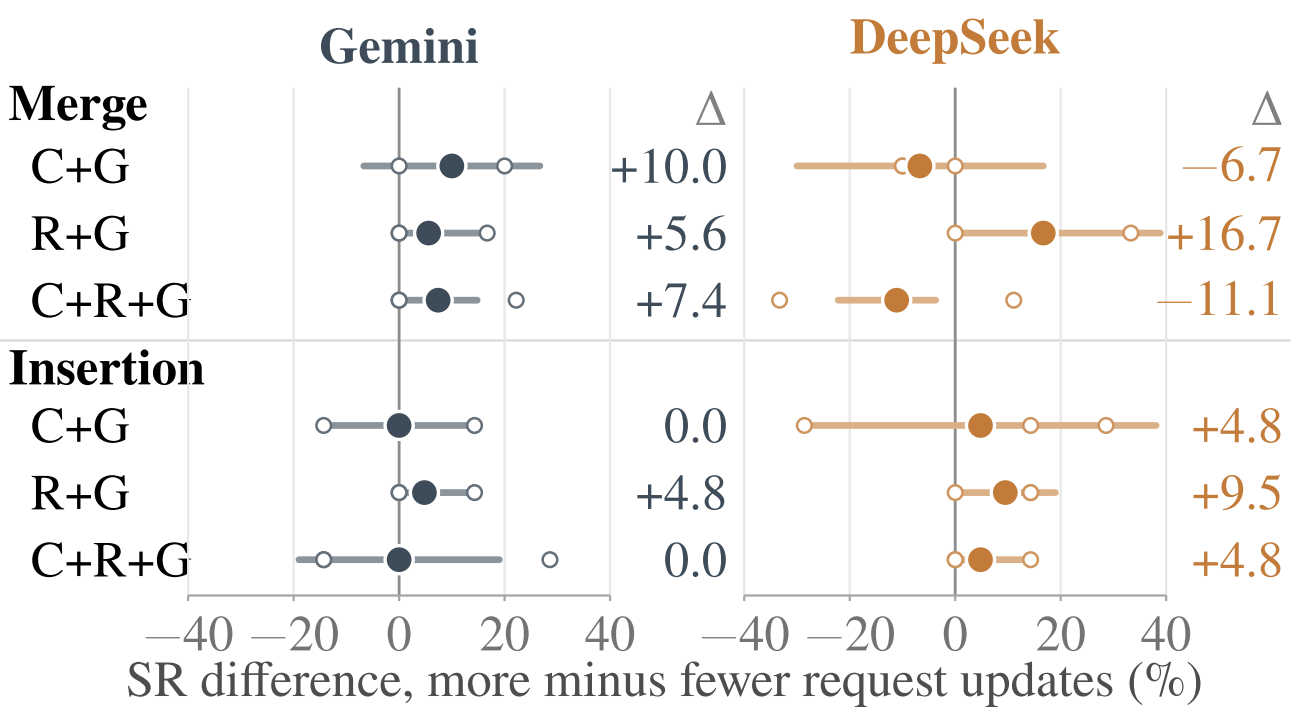
\begin{wrapfigure}{r}{.52\textwidth}
\centering
\resizebox{\linewidth}{!}{%
\begin{tikzpicture}[x=1cm,y=.70cm,font=\footnotesize,
  mean/.style={circle,fill=#1,draw=white,line width=.6pt,inner sep=0pt,minimum size=5pt},
  round/.style={circle,fill=white,draw=#1,line width=.45pt,inner sep=0pt,minimum size=2.6pt},
  meanval/.style={anchor=east,inner sep=0pt,text=#1}]
\newcommand{\cirow}[9]{%
  \draw[#1!60,line width=1.2pt,line cap=round] (#4,#2) -- (#5,#2);
  \foreach \r in {#6,#7,#8} {\node[round=#1!80] at (\r,#2) {};}
  \node[mean=#1] at (#3,#2) {};
  \node[meanval=#1] at (\valx,#2) {#9};}
\useasboundingbox (0,-5.08) rectangle (8.16,.62);
\node[anchor=west,font=\footnotesize\bfseries] at (0,0.00) {Merge};
\node[anchor=west,font=\footnotesize\bfseries] at (0,-2.32) {Insertion};
\draw[black!15,line width=.4pt] (0,-2.08) -- (8.16,-2.08);
\node[anchor=west] at (.14,-0.52) {C+G};
\node[anchor=west] at (.14,-1.12) {R+G};
\node[anchor=west] at (.14,-1.72) {C+R+G};
\node[anchor=west] at (.14,-2.84) {C+G};
\node[anchor=west] at (.14,-3.44) {R+G};
\node[anchor=west] at (.14,-4.04) {C+R+G};
\begin{scope}[shift={(2.580cm,0)},x=0.033cm]
\def\valx{62.4}
\node[anchor=south,font=\footnotesize\bfseries,text=resultslate] at (0,.2) {Gemini};
\node[anchor=east,inner sep=0pt,text=black!50] at (\valx,0.00) {$\Delta$};
\foreach \t in {-40,-20,20,40} {\draw[black!9,line width=.4pt] (\t,.18) -- (\t,-4.36);}
\draw[black!45,line width=.5pt] (0,.18) -- (0,-4.36);
\draw[black!35,line width=.4pt] (-40,-4.36) -- (40,-4.36);
\foreach \t/\l in {-40/\textminus40,-20/\textminus20,0/0,20/20,40/40} {\draw[black!35,line width=.4pt] (\t,-4.36) -- (\t,-4.41); \node[anchor=north,text=black!55,inner sep=2.5pt] at (\t,-4.41) {\l};}
\cirow{resultslate}{-0.52}{10.0}{-6.7}{26.7}{20.0}{10.0}{0.0}{+10.0}
\cirow{resultslate}{-1.12}{5.6}{0.0}{16.7}{16.7}{0.0}{0.0}{+5.6}
\cirow{resultslate}{-1.72}{7.4}{0.0}{14.8}{22.2}{0.0}{0.0}{+7.4}
\cirow{resultslate}{-2.84}{0.0}{-14.3}{14.3}{0.0}{14.3}{-14.3}{0.0}
\cirow{resultslate}{-3.44}{4.8}{0.0}{14.3}{0.0}{0.0}{14.3}{+4.8}
\cirow{resultslate}{-4.04}{0.0}{-19.0}{19.0}{-14.3}{28.6}{-14.3}{0.0}
\end{scope}
\begin{scope}[shift={(6.060cm,0)},x=0.033cm]
\def\valx{62.4}
\node[anchor=south,font=\footnotesize\bfseries,text=resultorange] at (0,.2) {DeepSeek};
\node[anchor=east,inner sep=0pt,text=black!50] at (\valx,0.00) {$\Delta$};
\foreach \t in {-40,-20,20,40} {\draw[black!9,line width=.4pt] (\t,.18) -- (\t,-4.36);}
\draw[black!45,line width=.5pt] (0,.18) -- (0,-4.36);
\draw[black!35,line width=.4pt] (-40,-4.36) -- (40,-4.36);
\foreach \t/\l in {-40/\textminus40,-20/\textminus20,0/0,20/20,40/40} {\draw[black!35,line width=.4pt] (\t,-4.36) -- (\t,-4.41); \node[anchor=north,text=black!55,inner sep=2.5pt] at (\t,-4.41) {\l};}
\cirow{resultorange}{-0.52}{-6.7}{-30.0}{16.7}{-10.0}{0.0}{-10.0}{\textminus6.7}
\cirow{resultorange}{-1.12}{16.7}{0.0}{38.9}{16.7}{0.0}{33.3}{+16.7}
\cirow{resultorange}{-1.72}{-11.1}{-22.2}{-3.7}{11.1}{-33.3}{-11.1}{\textminus11.1}
\cirow{resultorange}{-2.84}{4.8}{-28.6}{38.1}{28.6}{14.3}{-28.6}{+4.8}
\cirow{resultorange}{-3.44}{9.5}{0.0}{19.0}{14.3}{14.3}{0.0}{+9.5}
\cirow{resultorange}{-4.04}{4.8}{0.0}{14.3}{14.3}{0.0}{0.0}{+4.8}
\end{scope}
\node[anchor=north,text=black!70] at (4.24,-4.76) {SR difference, more minus fewer request updates (\%)};
\end{tikzpicture}%
}

\captionsetup{position=bottom,skip=11pt}
\caption{Within-task SR differences for more versus fewer request updates across 46 pairs. Filled markers and $\Delta$ give three-run means, bars show 95\% recording-level bootstrap intervals, and open markers show single-round differences. C, R, and G denote Clarify, Refresh, and Graft.}
\label{fig:request_updates}

\end{wrapfigure}

Figure~\ref{fig:request_updates} tests whether more request updates make tasks harder using 46 task pairs with three runs per version. Videos, websites, clarification rules, final goals, scheduled final update times, and step budgets remain fixed. Two requests are merged at the later scheduled time in 25 pairs, and an intermediate request is inserted in 21 pairs. Insertion leaves mean SR unchanged or higher in every comparison, while separating two requests can help or hurt. An extra request can prompt correction of an earlier choice, but can also be applied to the wrong target. Some differences reverse across repeated runs. More request updates therefore do not consistently make tasks harder.
\par
\endgroup

\FloatBarrier
\Needspace{6\baselineskip}
\subsection{{Failure Analysis}}
\label{sec:failure-analysis}
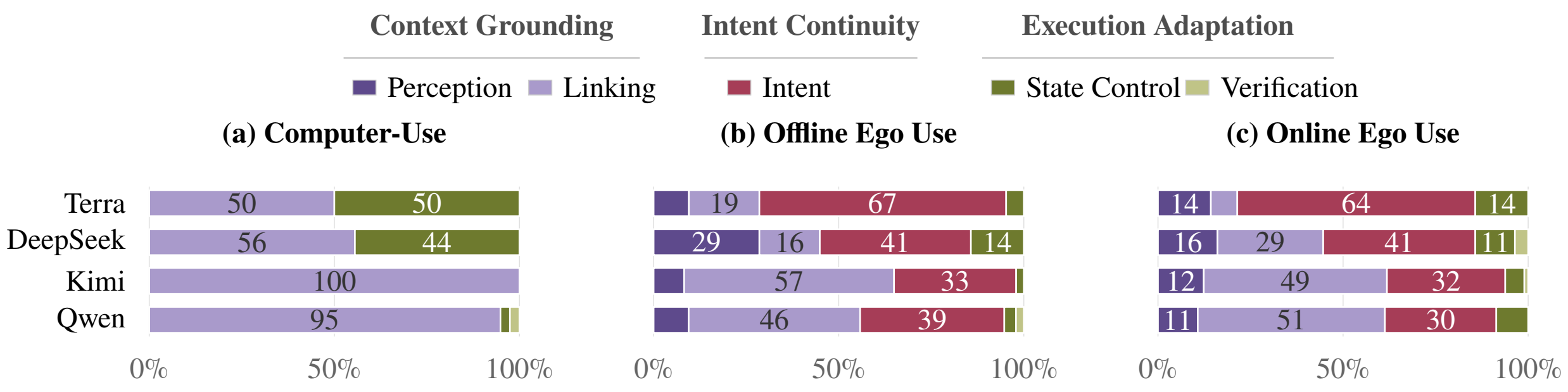
\begin{figure}[!htbp]
\centering
\resizebox{\linewidth}{!}{
\begin{tikzpicture}[x=1cm,y=1cm,font=\footnotesize]
\definecolor{failureP}{HTML}{5E4A8C}
\definecolor{failureL}{HTML}{A99ACB}
\definecolor{failureU}{HTML}{A63D57}
\definecolor{failureS}{HTML}{6E7A2E}
\definecolor{failureE}{HTML}{C0C487}
\definecolor{failureunattributed}{HTML}{D5D8DC}
\begin{scope}[xshift=1.90cm]
\node[anchor=south,font=\footnotesize\bfseries,text=black!70] at (2.85,1.62) {Context Grounding};
\draw[black!30,line width=.4pt] (1.25,1.57) -- (4.45,1.57);
\node[anchor=south,font=\footnotesize\bfseries,text=black!70] at (6.30,1.62) {Intent Continuity};
\draw[black!30,line width=.4pt] (5.15,1.57) -- (7.45,1.57);
\node[anchor=south,font=\footnotesize\bfseries,text=black!70] at (10.05,1.62) {Execution Adaptation};
\draw[black!30,line width=.4pt] (8.15,1.57) -- (11.95,1.57);
\filldraw[fill=failureP,draw=black!20,line width=.3pt] (1.35,1.17) rectangle ++(.25,.16);
\node[anchor=west,inner sep=2pt,text height=1.5ex,text depth=.3ex] at (1.65,1.25) {Perception};
\filldraw[fill=failureL,draw=black!20,line width=.3pt] (3.25,1.17) rectangle ++(.25,.16);
\node[anchor=west,inner sep=2pt,text height=1.5ex,text depth=.3ex] at (3.55,1.25) {Linking};
\filldraw[fill=failureU,draw=black!20,line width=.3pt] (5.40,1.17) rectangle ++(.25,.16);
\node[anchor=west,inner sep=2pt,text height=1.5ex,text depth=.3ex] at (5.70,1.25) {Intent};
\filldraw[fill=failureS,draw=black!20,line width=.3pt] (8.25,1.17) rectangle ++(.25,.16);
\node[anchor=west,inner sep=2pt,text height=1.5ex,text depth=.3ex] at (8.55,1.25) {State Control};
\filldraw[fill=failureE,draw=black!20,line width=.3pt] (10.35,1.17) rectangle ++(.25,.16);
\node[anchor=west,inner sep=2pt,text height=1.5ex,text depth=.3ex] at (10.65,1.25) {Verification};
\end{scope}
\begin{scope}[xshift=1.05cm]
\node[font=\footnotesize\bfseries] at (2.00,.73) {(a) Computer-Use};
\draw[black!12,line width=.3pt] (0.000,.18) -- (0.000,-1.440);
\node[anchor=north,text=black!60] at (0.000,-1.540) {0\%};
\draw[black!12,line width=.3pt] (2.000,.18) -- (2.000,-1.440);
\node[anchor=north,text=black!60] at (2.000,-1.540) {50\%};
\draw[black!12,line width=.3pt] (4.000,.18) -- (4.000,-1.440);
\node[anchor=north,text=black!60] at (4.000,-1.540) {100\%};
\node[anchor=east] at (-.12,0.000) {Terra};
\filldraw[fill=failureL,draw=white,line width=.5pt] (0.00000,-0.140) rectangle (2.00000,0.140);
\node[text=black!80,inner sep=0pt] at (1.00000,0.000) {50};
\filldraw[fill=failureS,draw=white,line width=.5pt] (2.00000,-0.140) rectangle (4.00000,0.140);
\node[text=white,inner sep=0pt] at (3.00000,0.000) {50};
\node[anchor=east] at (-.12,-0.420) {DeepSeek};
\filldraw[fill=failureL,draw=white,line width=.5pt] (0.00000,-0.560) rectangle (2.22222,-0.280);
\node[text=black!80,inner sep=0pt] at (1.11111,-0.420) {56};
\filldraw[fill=failureS,draw=white,line width=.5pt] (2.22222,-0.560) rectangle (4.00000,-0.280);
\node[text=white,inner sep=0pt] at (3.11111,-0.420) {44};
\node[anchor=east] at (-.12,-0.840) {Kimi};
\filldraw[fill=failureL,draw=white,line width=.5pt] (0.00000,-0.980) rectangle (4.00000,-0.700);
\node[text=black!80,inner sep=0pt] at (2.00000,-0.840) {100};
\node[anchor=east] at (-.12,-1.260) {Qwen};
\filldraw[fill=failureL,draw=white,line width=.5pt] (0.00000,-1.400) rectangle (3.79487,-1.120);
\node[text=black!80,inner sep=0pt] at (1.89744,-1.260) {95};
\filldraw[fill=failureS,draw=white,line width=.5pt] (3.79487,-1.400) rectangle (3.89744,-1.120);
\filldraw[fill=failureE,draw=white,line width=.5pt] (3.89744,-1.400) rectangle (4.00000,-1.120);
\end{scope}
\begin{scope}[xshift=6.50cm]
\node[font=\footnotesize\bfseries] at (2.00,.73) {(b) Offline Ego Use};
\draw[black!12,line width=.3pt] (0.000,.18) -- (0.000,-1.440);
\node[anchor=north,text=black!60] at (0.000,-1.540) {0\%};
\draw[black!12,line width=.3pt] (2.000,.18) -- (2.000,-1.440);
\node[anchor=north,text=black!60] at (2.000,-1.540) {50\%};
\draw[black!12,line width=.3pt] (4.000,.18) -- (4.000,-1.440);
\node[anchor=north,text=black!60] at (4.000,-1.540) {100\%};
\filldraw[fill=failureP,draw=white,line width=.5pt] (0.00000,-0.140) rectangle (0.38095,0.140);
\filldraw[fill=failureL,draw=white,line width=.5pt] (0.38095,-0.140) rectangle (1.14286,0.140);
\node[text=black!80,inner sep=0pt] at (0.76190,0.000) {19};
\filldraw[fill=failureU,draw=white,line width=.5pt] (1.14286,-0.140) rectangle (3.80952,0.140);
\node[text=white,inner sep=0pt] at (2.47619,0.000) {67};
\filldraw[fill=failureS,draw=white,line width=.5pt] (3.80952,-0.140) rectangle (4.00000,0.140);
\filldraw[fill=failureP,draw=white,line width=.5pt] (0.00000,-0.560) rectangle (1.14286,-0.280);
\node[text=white,inner sep=0pt] at (0.57143,-0.420) {29};
\filldraw[fill=failureL,draw=white,line width=.5pt] (1.14286,-0.560) rectangle (1.79592,-0.280);
\node[text=black!80,inner sep=0pt] at (1.46939,-0.420) {16};
\filldraw[fill=failureU,draw=white,line width=.5pt] (1.79592,-0.560) rectangle (3.42857,-0.280);
\node[text=white,inner sep=0pt] at (2.61224,-0.420) {41};
\filldraw[fill=failureS,draw=white,line width=.5pt] (3.42857,-0.560) rectangle (4.00000,-0.280);
\node[text=white,inner sep=0pt] at (3.71429,-0.420) {14};
\filldraw[fill=failureP,draw=white,line width=.5pt] (0.00000,-0.980) rectangle (0.32990,-0.700);
\filldraw[fill=failureL,draw=white,line width=.5pt] (0.32990,-0.980) rectangle (2.59794,-0.700);
\node[text=black!80,inner sep=0pt] at (1.46392,-0.840) {57};
\filldraw[fill=failureU,draw=white,line width=.5pt] (2.59794,-0.980) rectangle (3.91753,-0.700);
\node[text=white,inner sep=0pt] at (3.25773,-0.840) {33};
\filldraw[fill=failureS,draw=white,line width=.5pt] (3.91753,-0.980) rectangle (4.00000,-0.700);
\filldraw[fill=failureP,draw=white,line width=.5pt] (0.00000,-1.400) rectangle (0.37895,-1.120);
\filldraw[fill=failureL,draw=white,line width=.5pt] (0.37895,-1.400) rectangle (2.23158,-1.120);
\node[text=black!80,inner sep=0pt] at (1.30526,-1.260) {46};
\filldraw[fill=failureU,draw=white,line width=.5pt] (2.23158,-1.400) rectangle (3.78947,-1.120);
\node[text=white,inner sep=0pt] at (3.01053,-1.260) {39};
\filldraw[fill=failureS,draw=white,line width=.5pt] (3.78947,-1.400) rectangle (3.91579,-1.120);
\filldraw[fill=failureE,draw=white,line width=.5pt] (3.91579,-1.400) rectangle (4.00000,-1.120);
\end{scope}
\begin{scope}[xshift=11.95cm]
\node[font=\footnotesize\bfseries] at (2.00,.73) {(c) Online Ego Use};
\draw[black!12,line width=.3pt] (0.000,.18) -- (0.000,-1.440);
\node[anchor=north,text=black!60] at (0.000,-1.540) {0\%};
\draw[black!12,line width=.3pt] (2.000,.18) -- (2.000,-1.440);
\node[anchor=north,text=black!60] at (2.000,-1.540) {50\%};
\draw[black!12,line width=.3pt] (4.000,.18) -- (4.000,-1.440);
\node[anchor=north,text=black!60] at (4.000,-1.540) {100\%};
\filldraw[fill=failureP,draw=white,line width=.5pt] (0.00000,-0.140) rectangle (0.57143,0.140);
\node[text=white,inner sep=0pt] at (0.28571,0.000) {14};
\filldraw[fill=failureL,draw=white,line width=.5pt] (0.57143,-0.140) rectangle (0.85714,0.140);
\filldraw[fill=failureU,draw=white,line width=.5pt] (0.85714,-0.140) rectangle (3.42857,0.140);
\node[text=white,inner sep=0pt] at (2.14286,0.000) {64};
\filldraw[fill=failureS,draw=white,line width=.5pt] (3.42857,-0.140) rectangle (4.00000,0.140);
\node[text=white,inner sep=0pt] at (3.71429,0.000) {14};
\filldraw[fill=failureP,draw=white,line width=.5pt] (0.00000,-0.560) rectangle (0.64286,-0.280);
\node[text=white,inner sep=0pt] at (0.32143,-0.420) {16};
\filldraw[fill=failureL,draw=white,line width=.5pt] (0.64286,-0.560) rectangle (1.78571,-0.280);
\node[text=black!80,inner sep=0pt] at (1.21429,-0.420) {29};
\filldraw[fill=failureU,draw=white,line width=.5pt] (1.78571,-0.560) rectangle (3.42857,-0.280);
\node[text=white,inner sep=0pt] at (2.60714,-0.420) {41};
\filldraw[fill=failureS,draw=white,line width=.5pt] (3.42857,-0.560) rectangle (3.85714,-0.280);
\node[text=white,inner sep=0pt] at (3.64286,-0.420) {11};
\filldraw[fill=failureE,draw=white,line width=.5pt] (3.85714,-0.560) rectangle (4.00000,-0.280);
\filldraw[fill=failureP,draw=white,line width=.5pt] (0.00000,-0.980) rectangle (0.49485,-0.700);
\node[text=white,inner sep=0pt] at (0.24742,-0.840) {12};
\filldraw[fill=failureL,draw=white,line width=.5pt] (0.49485,-0.980) rectangle (2.47423,-0.700);
\node[text=black!80,inner sep=0pt] at (1.48454,-0.840) {49};
\filldraw[fill=failureU,draw=white,line width=.5pt] (2.47423,-0.980) rectangle (3.75258,-0.700);
\node[text=white,inner sep=0pt] at (3.11340,-0.840) {32};
\filldraw[fill=failureS,draw=white,line width=.5pt] (3.75258,-0.980) rectangle (3.95876,-0.700);
\filldraw[fill=failureE,draw=white,line width=.5pt] (3.95876,-0.980) rectangle (4.00000,-0.700);
\filldraw[fill=failureP,draw=white,line width=.5pt] (0.00000,-1.400) rectangle (0.43011,-1.120);
\node[text=white,inner sep=0pt] at (0.21505,-1.260) {11};
\filldraw[fill=failureL,draw=white,line width=.5pt] (0.43011,-1.400) rectangle (2.45161,-1.120);
\node[text=black!80,inner sep=0pt] at (1.44086,-1.260) {51};
\filldraw[fill=failureU,draw=white,line width=.5pt] (2.45161,-1.400) rectangle (3.65591,-1.120);
\node[text=white,inner sep=0pt] at (3.05376,-1.260) {30};
\filldraw[fill=failureS,draw=white,line width=.5pt] (3.65591,-1.400) rectangle (4.00000,-1.120);
\end{scope}
\end{tikzpicture}}
\caption{{Failure composition across three interaction settings. Legend headings show primary challenge associations. Segments show percentages within each model-setting group.}}
\label{fig:failure-attribution}
\end{figure}

{Figure~\ref{fig:failure-attribution} compares 604 eligible failed trajectories from Terra, DeepSeek, Kimi, and qwen3.5-27b across the three settings. gpt-6-luna analyzes the trajectories, and a human reviewer assigns the cases without a definite model judgment using task and trajectory summaries. The supplementary material details the trajectory evaluation procedure.}

\paragraph{{Context Grounding.}}
{Agents fail to match the user's description or visual reference to a suitable website item. Linking accounts for 90.2\% of Computer-Use failures and leads for Kimi and qwen3.5-27b in both Ego Use settings, covering 50.0\% of their Online failures. Perception errors occur when agents act on attributes contradicted by already delivered frames.}

\paragraph{{Intent Continuity.}}
{Agents may commit to an unresolved reference or continue pursuing a target after the request changes. These errors concern which target to pursue, rather than how to edit the website. Intent is the leading category for Terra and DeepSeek in both Ego Use settings and accounts for 45.7\% of their Online failures.}

\paragraph{{Execution Adaptation.}}
{Agents may find a correct replacement but fail to remove an obsolete selection. State Control and Verification account for 9.2\% of Online failures. Revision opportunities also depend on the state reached before delivery. Across six main Online models, 14.6\% of delivered Graft and Refresh updates arrive when the website satisfies the initial request but needs changes to meet the revised one. The supplementary material analyzes this subset.}

\section{Conclusion}
We introduce \textsc{EgoUseBench}, a benchmark of 693 tasks across Computer-Use, Offline Ego Use, and Online Ego Use. Our analysis reveals two main difficulties in Online Ego Use. Agents struggle to identify the user's intended target from first-person observations and to update website state as requirements change. In particular, they sometimes find a valid new option while retaining earlier choices that no longer satisfy the request. These findings point to a broader challenge in Ego Use. Agents struggle to keep their website actions aligned with the user's current intent and evolving physical context. \textsc{EgoUseBench} provides a controlled testbed for studying this alignment. More broadly, Ego Use offers a research direction for bringing first-person agents into a wider range of everyday settings, where what people see and do guides computer tasks and agents adapt as circumstances change.

{\small
\setlength{\bibsep}{1pt plus .5pt}
\bibliographystyle{unsrtnat}
\bibliography{main}
}

\newsavebox{\appendixdisplaybox}
\newenvironment{appendixdisplay}[2]{%
  \par\begin{lrbox}{\appendixdisplaybox}%
  \begin{minipage}{#2}\captionsetup{type=#1,justification=raggedright,singlelinecheck=false}%
}{%
  \end{minipage}\end{lrbox}%
  \ifinner\else
    \Needspace{\dimexpr\ht\appendixdisplaybox+\dp\appendixdisplaybox+\intextsep\relax}%
  \fi
  \addvspace{\intextsep}\noindent\makebox[\linewidth][c]{\usebox{\appendixdisplaybox}}\par
  \addvspace{\intextsep}%
}
\renewenvironment{table}[1][]{\begin{appendixdisplay}{table}{\columnwidth}}{\end{appendixdisplay}}
\renewenvironment{figure}[1][]{\begin{appendixdisplay}{figure}{\columnwidth}}{\end{appendixdisplay}}
\renewenvironment{table*}[1][]{\begin{appendixdisplay}{table}{\textwidth}}{\end{appendixdisplay}}
\renewenvironment{figure*}[1][]{\begin{appendixdisplay}{figure}{\textwidth}}{\end{appendixdisplay}}
\newenvironment{appendixcompacttable}{\begin{appendixdisplay}{table}{.76\linewidth}}{\end{appendixdisplay}}
\newenvironment{appendixcompactfigure}{\begin{appendixdisplay}{figure}{.76\linewidth}}{\end{appendixdisplay}}

\newenvironment{appendixpanels}[1][.49\linewidth]{%
  \def\appendixpanelwidth{#1}%
  \par\medskip\noindent\begin{minipage}{\linewidth}%
  \begin{minipage}[t]{\appendixpanelwidth}\vspace{0pt}%
}{%
  \end{minipage}\end{minipage}\par\medskip%
}
\newcommand{\appendixnextpanel}{%
  \end{minipage}\hfill\begin{minipage}[t]{\appendixpanelwidth}\vspace{0pt}%
}
\newenvironment{appendixlongtable}[1]{%
  \par%
  \small
  \renewcommand{\arraystretch}{1.12}%
  \setlength{\LTleft}{0pt plus 1fil}%
  \setlength{\LTright}{0pt plus 1fil}%
  \setlength{\LTcapwidth}{\linewidth}%
  \setlength{\LTpre}{\intextsep}%
  \setlength{\LTpost}{\intextsep}%
  \captionsetup{justification=raggedright,singlelinecheck=false}%
  \begin{longtable}{@{\extracolsep{\fill}}#1@{}}%
}{\end{longtable}}
\newcommand{\appendixcontinued}[1]{%
  \multicolumn{#1}{@{}l}{\textit{Table~\thetable\ (continued)}}\\[2pt]%
}
\newenvironment{appendixcostpanel}{%
  \begin{minipage}[t]{\linewidth}%
  \footnotesize
  \setlength{\tabcolsep}{1.5pt}%
  \begin{tabular*}{\linewidth}{@{\extracolsep{\fill}}lrrrr@{}}%
  \toprule
  Measure & Offline & Online & $\Delta$ & 95\% CI\\
  \midrule
}{\bottomrule\end{tabular*}\end{minipage}}
\newgeometry{textwidth=7in,top=.88in,bottom=.88in,footskip=28pt}
\appendix
\raggedbottom

\setcounter{table}{0}
\setcounter{figure}{0}
\setcounter{equation}{0}
\renewcommand{\thetable}{S\arabic{table}}
\renewcommand{\thefigure}{S\arabic{figure}}
\renewcommand{\theequation}{S\arabic{equation}}
\renewcommand{\theHtable}{appendix.\arabic{table}}
\renewcommand{\theHfigure}{appendix.\arabic{figure}}
\renewcommand{\theHequation}{appendix.\arabic{equation}}
\setlength{\intextsep}{6pt plus 1pt minus 1pt}

\pdfbookmark[0]{Appendix Contents}{appendix-overview}
\etocdepthtag.toc{appendix}
\begingroup
\small
\setlength{\parskip}{0pt}
\hypersetup{linktoc=all,linkcolor=black}
\etocsetnexttocdepth{subsection}
\etocsettagdepth{main}{none}
\etocsettagdepth{appendix}{subsection}
\etocsetstyle{section}{}{\addvspace{7.5pt}}
  {\noindent{\sffamily\bfseries%
    \makebox[2em][l]{\etocnumber}\etocname\hfill\etocpage}\par\nobreak\vspace{1pt}\nobreak}{}
\etocsetstyle{subsection}{}{\nopagebreak[4]}
  {\noindent\hspace{2em}\makebox[2.6em][l]{\etocnumber}\etocname
   \enspace{\color{PreprintRule}\dotfill}\enspace\etocpage\par}{}
\etocsettocstyle{\section*{Appendix Contents}}{\par\addvspace{\medskipamount}}
\tableofcontents
\endgroup
\newpage

\section{Task Construction}
\label{app:task-checks}

\subsection{Source Data}
\label{app:source-coverage}
Table~\ref{tab:source-coverage} lists the sources of the 200 Online Ego Use tasks used in the matched comparison. These tasks cover 82 shopping, 117 tutorial retrieval, and one place bookmarking request, with one task per original recording. For Ego2Web and EgoSchema clips, we identify the recording by its Ego4D parent.

The composite set adds 93 Online Ego Use tasks from 93 recordings: 63 shopping and 30 tutorial retrieval tasks. The two Online sets share 79 recordings, giving 214 distinct recordings. The Computer-Use and Offline Ego Use settings each add 200 matched task versions, bringing the benchmark to 693 tasks.

\begin{appendixcompacttable}
\centering\small
\caption{Task counts by source dataset and task type in the main set.}
\label{tab:source-coverage}
\renewcommand{\arraystretch}{1.12}
\begin{tabularx}{\columnwidth}{@{}Xrrrr@{}}
\toprule
Source & Graft & Refresh & Clarify & Total\\
\midrule
Ego2Web & 112 & 4 & 16 & 132\\
CaptainCook4D & 23 & 4 & 1 & 28\\
HD-EPIC & 5 & 13 & 10 & 28\\
EgoOops & 1 & 4 & 1 & 6\\
HoloAssist & 0 & 4 & 0 & 4\\
EgoLife & 1 & 0 & 0 & 1\\
EgoSchema & 0 & 1 & 0 & 1\\
\midrule
Total & 142 & 30 & 28 & 200\\
\bottomrule
\end{tabularx}
\end{appendixcompacttable}

The website catalogs contain 11,660 product entities, 16,389 video entities, and 289 place entities. BM25 retrieves candidates from public listing fields in these shared catalogs. Accepted alternatives are not marked in search results.

\subsection{Task Annotation}
\paragraph{Visual grounding.}
Annotations identify the physical referent and the evidence needed to interpret each request. Object identity, attributes, actions, and temporal relations are specified at the level supported by the released frames. Exact commercial identities require distinguishing evidence beyond generic color or shape. Source events and authored user requests are annotated separately.

Graft follow-ups express plausible changes to the wearer's requirements. Refresh annotations identify whether later visual evidence changes the acceptable choices. Clarify replies distinguish the intended physical referent without naming a website entity. The complete replay window is checked for premature disclosure of later evidence.

\paragraph{Website matching.}
Candidate review considers plausible alternatives in the public catalog. Accepted entities require positive evidence of relevance or compatibility. Missing specifications are treated as unresolved rather than as evidence of incompatibility. Tutorial relevance is established from public descriptions, chapters, subtitles, or captured previews. Brand and model constraints are retained only when supported by the request or visual evidence.

Rendered pages and agent-readable observations use the same website facts. Source annotations and answer labels are excluded from both. Tasks reference website entities, and the evaluator accepts all verified alternatives satisfying the request.

\paragraph{Validation.}
An independent Coding Agent first records a solution without access to the proposed answers, then reviews answer coverage and information shortcuts. Development runs with Qwen3.5-Flash, including text-only shortcut tests, informed task revisions before model evaluation. Revisions are checked against the source recording and public website evidence. Each reported comparison uses its fixed task and website version.

\subsection{Human Quality Review}
\label{app:human-quality}
Two independent human reviewers assessed the same 40 tasks from 40 distinct recordings: 14 Graft, 10 Refresh, 8 Clarify, and 8 composite tasks. Sampling was stratified by task type without considering model outcomes. For each task, the review interface presented the requests in temporal order, sampled frames from the complete reviewed window, the standard clarification reply where applicable, all registered acceptable outcomes with quantities and fields, and the corresponding website candidates and details. Model results and previous review comments were not shown.

Reviewers checked whether the visual evidence supported the request, whether the registered answers were valid and covered reasonable alternatives, and whether the updates and clarification were natural and meaningful. Each task received one of three judgments: \emph{Acceptable}, \emph{Problem}, or \emph{Uncertain}. Problem and uncertain judgments required a short explanation.

Table~\ref{tab:human-quality} breaks down the joint acceptance rate reported in the main paper by task type. Each reviewer accepted 95\% of the sample, and 90\% of the sample was accepted by both. Reviewer~1 marked one Graft task as problematic and another as uncertain. Reviewer~2 marked one Clarify task and one composite task as uncertain.

\begin{appendixcompacttable}
\centering\small
\caption{Human quality-review counts on the same 40 tasks. Reviewer columns count tasks judged acceptable, and Both counts tasks accepted by both reviewers.}
\label{tab:human-quality}
\renewcommand{\arraystretch}{1.12}
\begin{tabularx}{\columnwidth}{@{}Xrrrr@{}}
\toprule
Task type & $n$ & Reviewer 1 & Reviewer 2 & Both\\
\midrule
Graft & 14 & 12 & 14 & 12\\
Refresh & 10 & 10 & 10 & 10\\
Clarify & 8 & 8 & 7 & 7\\
Composite & 8 & 8 & 7 & 7\\
\midrule
Total & 40 & 38 & 38 & 36\\
\bottomrule
\end{tabularx}
\end{appendixcompacttable}

\subsection{Evaluator Tests}
Evaluator tests apply controlled modifications to accepted states. Removing required content, adding unrelated content, changing a required quantity, or retaining an obsolete choice produces a failure unless the result matches another accepted alternative. Verified alternatives remain valid under entity ordering and registered packaging equivalences.

Scoring uses the terminal website state. Corrected intermediate errors do not affect success, whereas errors remaining at termination do. A clarification question and a separate submit action are not scoring requirements. Runs without terminal evidence are excluded from scored outcomes.

\section{Evaluation Protocol}
\label{app:experiment-details}

\subsection{Action Interface}
\label{app:implementation}
Each step takes one JSON object with an \texttt{action} name and a \texttt{parameters} object. For example, a click uses \texttt{action: click} with \texttt{bid} in its parameters. The \texttt{bid} identifies an element in the current accessibility tree. The interface exposes the following actions on the local task websites:

\begin{description}
\setlength{\itemsep}{0pt}
\item[Elements.] \texttt{click} takes \texttt{bid} and optionally \texttt{button} and \texttt{modifiers}. \texttt{fill} takes \texttt{bid} and \texttt{value}, \texttt{press} takes \texttt{bid} and \texttt{key\_comb}, and \texttt{select\_option} takes \texttt{bid} and \texttt{options}. The last parameter accepts one option or a list. \texttt{clear} and \texttt{hover} take only \texttt{bid}.
\item[Browser.] \texttt{scroll} takes \texttt{delta\_x} and \texttt{delta\_y}, \texttt{goto} takes \texttt{url}, and \texttt{tab\_focus} takes \texttt{index}. \texttt{go\_back}, \texttt{go\_forward}, \texttt{new\_tab}, and \texttt{tab\_close} take no parameters. \texttt{noop} takes an optional \texttt{wait\_ms} from 0 to 1000, with 1000 as its default.
\item[Task control.] \texttt{ask\_user} takes a question string in \texttt{question}. \texttt{wait} takes no parameters and advances replay without a website action. \texttt{done} takes optional \texttt{text} and declares completion. For a pending Graft follow-up, the delivery rules may resume execution.
\end{description}

The first Clarify question returns the task's fixed reply. In Graft and Refresh, it returns \emph{Just what I asked for.} Repeated questions return \emph{I've already answered.} Questions consume a decision step, and the reply is available before the next decision. Computer-Use disables questions because its request already states the resolved requirements.

\subsection{Image Rendering}
\label{app:render-protocol}
{The default layout groups four time-stamped frames into a JPEG tile at quality 85. Five main-set tasks instead send each source frame as a separate image, and seven use a maximum tile-edge length. Table~\ref{tab:render-limits} lists these settings. Size-limited tiles are resized proportionally with Lanczos resampling and encoded at quality 85. These settings are fixed by task across models, Ego Use settings, and controlled conditions. They preserve the sampled frames, their order and timestamps, and their scheduled delivery. Webpage screenshots are unchanged. Compressed history retains every other frame within older batches and uses the same per-task layout.}

\begin{appendixcompacttable}
\centering\small
\renewcommand{\arraystretch}{1.12}
\caption{{Task-specific image layouts and size limits in the main set. Windows are continuous source-video intervals, and frame counts refer to the 1\,fps baseline. Limits give the maximum tile edge in pixels.}}
\label{tab:render-limits}
\begin{tabularx}{\columnwidth}{@{}>{\raggedright\arraybackslash}Xrrr@{}}
\toprule
Task & Window (s) & Frames & Limit\\
\midrule
\multicolumn{4}{@{}l}{{Four-frame tiles with size limits}}\\
Tea strainer (0678) & 18--128 & 111 & 1280\\
Skull mask (0622) & 5--164 & 160 & 1024\\
Cutlery (1308) & 2548--2607 & 60 & 1280\\
Chocolate dessert (0531) & 782--845 & 64 & 1280\\
{Spice refill (1601)} & {504--565} & {62} & {1536}\\
{Pan utensil (retain-08)} & {709.17--741.80} & {33} & {1280}\\
{Carrot soup (0509)} & {13--76} & {64} & {1280}\\
\midrule
\multicolumn{4}{@{}l}{{Individual source frames}}\\
{Kitchen items (2802)} & {20--87.5} & {68} & {n/a}\\
{Workbench tool (2702)} & {0--78} & {79} & {n/a}\\
{Bedding guide (c-131)} & {61--179.5} & {119} & {n/a}\\
{Surface cleaning (c-64)} & {10--84} & {75} & {n/a}\\
{Cooking and hair care (r-68)} & {20--179} & {160} & {n/a}\\
\bottomrule
\end{tabularx}
\end{appendixcompacttable}

\subsection{Input Variants}
Computer-Use requests state the final requirements without website answer identifiers or action sequences. Offline Ego Use provides the same reviewed recording window before execution. Graft utterances preserve their wording, order, and source-time references. Refresh retains the original request. Both Ego Use settings preserve the ambiguous Clarify request and its question-triggered reply. All settings use the same website, initial state, and acceptable final outcomes.

The Cued + facts condition uses descriptions verified against every frame of the change batch for all 30 Refresh tasks. Descriptions concern visible objects, attributes, and actions, without future evidence, website answers, or instructions to revise a selection. The three Cue conditions use the same images and requests, with 15 Replace and 15 Retain tasks selected before model evaluation. Descriptions are delivered with the change batch and do not resume a terminated Refresh episode. The comparison includes the effects of both visual description and textual attention cues.

\subsection{Statistical Analysis}
Overall SR gives each task equal weight. Paired comparisons use tasks scored in all required conditions, with the same tasks and weights on both sides. We average outcomes within original recordings and then across recordings. Confidence intervals reflect variation across recordings, using 2,000 recording-level bootstrap resamples with seed zero. Each task has one run per condition unless otherwise specified. The request-update count comparison averages three runs per version before resampling.

Following the conditional comparison in InterventionBench~\cite{lee2026interventionbench}, adaptation rate (AR) measures Online success among tasks completed Offline. Let $\mathcal{P}$ denote complete, valid Offline and Online pairs for Graft or Refresh. For pair $i$, let $o_i,n_i\in\{0,1\}$ indicate Offline and Online success. Then
\begin{equation}
 \mathrm{AR}=\frac{\sum_{i\in\mathcal{P}}o_i n_i}{\sum_{i\in\mathcal{P}}o_i}.
 \label{eq:adaptation-rate}
\end{equation}
Pairs are equally weighted. A zero denominator is reported as n/a. Since the Offline-success subset depends on the model, AR is reported with both settings' SR and the paired outcome counts. Clarify is analyzed separately as referent disambiguation.

\section{Supplementary Results}
\label{app:additional-results}

\subsection{Online Completion}
\label{app:online-coverage}
Table~\ref{tab:online-counts} gives the success counts behind Table~\ref{tab:main}, with overall confidence intervals and the additional UI-TARS baseline. Each model has scored outcomes on all 200 Online tasks. All 600 qwen3.5-27b outcomes use Qianwen, with one outcome per task and setting. Earlier FP8 runs are retained in the experiment archives and excluded from these results.

\begin{appendixcompacttable}

\centering\small
\caption{Online successes out of scored tasks and 95\% recording-level bootstrap intervals for overall SR.}
\label{tab:online-counts}
\renewcommand{\arraystretch}{1.12}
\begin{tabular*}{\linewidth}{@{\extracolsep{\fill}}lrrrr@{}}
\toprule
Model & Graft & Refresh & Clarify & Overall 95\% CI (\%)\\
\midrule
GPT-6.1 Terra & 126/142 & 25/30 & 23/28 & [82.0, 91.5]\\
Gemini 3.8 Flash & 108/142 & 23/30 & 21/28 & [70.0, 81.5]\\
Doubao Seed 2.1 Pro & 111/142 & 19/30 & 20/28 & [69.0, 80.5]\\
DeepSeek-V4.1-Flash-1017 & 98/142 & 13/30 & 14/28 & [55.5, 69.0]\\
Kimi K2.6 & 59/142 & 8/30 & 11/28 & [32.0, 46.0]\\
qwen3.5-27b & 55/142 & 11/30 & 16/28 & [34.5, 48.0]\\
UI-TARS-1.5-7B & 0/142 & 0/30 & 0/28 & [0.0, 0.0]\\
\bottomrule\end{tabular*}
\end{appendixcompacttable}

\paragraph{UI-TARS.}
Under the shared JSON action interface, UI-TARS obtains zero successes in each of the three settings in Table~\ref{tab:service-settings}. Its traces contain unsupported action names, invalid element identifiers, and malformed responses, including in Computer-Use. All 600 episodes have scored terminal states. Of these, 36 use JSON-object decoding without an output cap, and 564 use a 512-token output cap without API-enforced JSON decoding. Both configurations request the same action format. These results concern the shared interface rather than a native UI-TARS adapter. {These episodes are reported separately from the failure analysis in the main paper.}

\paragraph{{Delivery modes.}}
{All 142 Graft follow-ups reach Gemini. SR is 77.6\% for delivery during execution ($n=116$) and 69.2\% for delivery after standby ($n=26$).}

\paragraph{{Update-time state.}}
{Table~\ref{tab:update-state} covers 172 Graft and Refresh episodes per model. It groups runs that received an update by the website state at delivery and lists undelivered updates separately. Revision required denotes a state satisfying the initial but not the final request. Initial incomplete includes unfinished or incorrect initial selections. No selection held also covers runs in which the agent removed an earlier selection before delivery. Each run is counted in one group.}

\begin{table*}\centering

\caption{Final successes over runs by website state at update delivery. No delivery reports only the number of runs that did not receive the required update. Task coverage follows Table~\ref{tab:online-counts}.}
\label{tab:update-state}\small
\renewcommand{\arraystretch}{1.12}
\begin{tabular*}{\textwidth}{@{\extracolsep{\fill}}lrrrrr@{}}\toprule
Model & No selection held & Revision required & Final goal already met & Initial incomplete & No delivery\\\midrule
Terra & 117/135 & 28/31 & 4/4 & 2/2 & 0\\
Gemini & 107/139 & 19/26 & 4/4 & 0/2 & 1\\
Doubao & 100/131 & 24/30 & 4/4 & 2/5 & 2\\
DeepSeek & 88/136 & 17/23 & 3/3 & 2/7 & 3\\
Kimi & 53/132 & 8/17 & 3/4 & 3/15 & 4\\
qwen3.5-27b & 52/128 & 11/22 & 3/4 & 0/15 & 3\\
UI-TARS & 0/172 & 0/0 & 0/0 & 0/0 & 0\\
\bottomrule\end{tabular*}\end{table*}

\subsection{Interaction Settings}
Table~\ref{tab:settings} extends Table~\ref{tab:matched-settings-sr} with paired differences and adaptation rates. Each of the 200 triplets per model uses the same task, starting website state, and final-state criterion. Table~\ref{tab:setting-pairs} shows which tasks succeed in Offline, Online, or both, and Table~\ref{tab:success-differences} gives paired SR differences.

Across these six models, equally weighted mean SR is 86.1\% in Computer-Use, 64.2\% in Offline Ego Use, and 63.4\% in Online Ego Use. The relative decreases in Online Ego Use are 26.3\% from Computer-Use and 1.2\% from Offline Ego Use, calculated from unrounded means.

Similar success rates can hide changes in which tasks are completed. For example, qwen3.5-27b succeeds on 83 tasks in Offline and 82 in Online, but its outcome changes on 53 of the 200 tasks. Figure~\ref{fig:setting-transitions} shows the 27 losses and 26 gains behind that near-tie. Kimi shows a similar pattern, with 23 losses and 20 gains. On Graft and Refresh tasks, both models retain fewer than 70\% of their Offline successes.

\begin{appendixcompactfigure}
\centering
\scalebox{1.18}{%
\begin{tikzpicture}[x=1cm,y=1cm,font=\scriptsize]
  \useasboundingbox (-.5,-.65) rectangle (7.7,4.1);
  \node[anchor=west,font=\small\bfseries] at (-.5,3.9) {Offline to Online};
  \node[text=resultorange!85!black] at (2.95,3.45) {Lost};
  \node[text=resultteal!85!black] at (5.05,3.45) {Gained};
  \node[text=black!65] at (7.05,3.45) {Net $\Delta$SR (\%)};
  \foreach \t/\label in {-30/\textminus30,-15/\textminus15,0/0,15/15,30/30} {
    \draw[black!10,line width=.4pt] ({4+.07*\t},.2) -- ({4+.07*\t},3.2);
    \draw[black!40,line width=.4pt] ({4+.07*\t},.2) -- ({4+.07*\t},.14);
    \node[anchor=north,text=black!55,inner sep=2pt] at ({4+.07*\t},.13) {\label};
  }
  \draw[black!40,line width=.4pt] (1.9,.2) -- (6.1,.2);
  \draw[black!55,line width=.6pt] (4,.2) -- (4,3.2);
  \node[anchor=north,text=black!60] at (4,-.28) {Tasks};
  \foreach \name/\y/\lost/\gained/\delta in {
    Terra/3/5/11/{+3.0},
    Gemini/2.5/20/19/{\textminus0.5},
    Doubao/2/21/18/{\textminus1.5},
    DeepSeek/1.5/27/20/{\textminus3.5},
    Kimi/1/23/20/{\textminus1.5},
    qwen3.5-27b/0.5/27/26/{\textminus0.5}} {
    \node[anchor=east,inner sep=0pt] at (1.45,\y) {\name};
    \fill[resultorange!80] ({4-.07*\lost},{\y-.13}) rectangle (4,{\y+.13});
    \fill[resultteal!85] (4,{\y-.13}) rectangle ({4+.07*\gained},{\y+.13});
    \node[anchor=east,inner sep=0pt,text=resultorange!85!black] at ({4-.07*\lost-.08},\y) {\textminus\lost};
    \node[anchor=west,inner sep=0pt,text=resultteal!85!black] at ({4+.07*\gained+.08},\y) {+\gained};
    \ifnum\gained>\lost
      \node[text=resultteal!85!black] at (7.05,\y) {\textbf{\delta}};
    \else
      \node[text=resultorange!85!black] at (7.05,\y) {\textbf{\delta}};
    \fi
  }
\end{tikzpicture}%
}
\caption{Paired Offline and Online outcomes on 200 tasks per model. Lost denotes Offline success and Online failure, and Gained denotes the reverse. Bars show task counts. Net $\Delta$SR (\%) is Online minus Offline.}
\label{fig:setting-transitions}
\end{appendixcompactfigure}

\begin{table}

\centering
\caption{Matched comparison across three interaction settings for all six main models. SR and AR are reported in \%. $\Delta_1$ (\%) is Offline minus Computer-Use and $\Delta_2$ (\%) is Online minus Offline. Shading marks the Online condition. Each model has 200 complete task triplets (142 Graft, 15 Replace, 15 Retain, and 28 Clarify), from 200 recordings.}
\label{tab:settings}
\renewcommand{\MatchedOnlineColor}{\rowcolor{resultshade}}
\newcommand{\MatchedSR}[9]{#2 & #3 & #5 & #6 & #7\\}
\small
\renewcommand{\arraystretch}{1.03}
\setlength{\tabcolsep}{2.5pt}
\begin{appendixpanels}
\begin{tabularx}{\columnwidth}{@{}p{.34\columnwidth}*{4}{>{\raggedleft\arraybackslash}X}@{}}
\toprule
& & \multicolumn{2}{c}{\textbf{Refresh}} & \\
\cmidrule(lr){3-4}
\textbf{Setting} & \textbf{Graft} & Replace & Retain & \textbf{Clarify}\\
\midrule
\rowcolor{resultheader}\multicolumn{5}{l}{\textbf{GPT-6.1 Terra}}\\
\MatchedSRTerra
$\Delta_1$ & -3.5 & -6.7 & -40.0 & -28.6\\
$\Delta_2$ & 0.0 & 0.0 & +20.0 & +10.7\\
AR & 96.8 & 92.3 & 100.0 & \textcolor{black!40}{n/a}\\
\addlinespace
\rowcolor{resultheader}\multicolumn{5}{l}{\textbf{Gemini 3.8 Flash}}\\
\MatchedSRGemini
$\Delta_1$ & -8.5 & -13.3 & -26.7 & -28.6\\
$\Delta_2$ & -2.1 & -6.7 & +13.3 & +3.6\\
AR & 85.6 & 91.7 & 100.0 & \textcolor{black!40}{n/a}\\
\addlinespace
\rowcolor{resultheader}\multicolumn{5}{l}{\textbf{Doubao Seed 2.1 Pro}}\\
\MatchedSRDoubao
$\Delta_1$ & -9.2 & -6.7 & -26.7 & -46.4\\
$\Delta_2$ & -3.5 & -33.3 & +6.7 & +21.4\\
AR & 89.7 & 53.8 & 100.0 & \textcolor{black!40}{n/a}\\
\bottomrule
\end{tabularx}
\appendixnextpanel
\begin{tabularx}{\columnwidth}{@{}p{.34\columnwidth}*{4}{>{\raggedleft\arraybackslash}X}@{}}
\toprule
& & \multicolumn{2}{c}{\textbf{Refresh}} & \\
\cmidrule(lr){3-4}
\textbf{Setting} & \textbf{Graft} & Replace & Retain & \textbf{Clarify}\\
\midrule
\rowcolor{resultheader}\multicolumn{5}{l}{\textbf{DeepSeek-V4.1-Flash-1017}}\\
\MatchedSRDeepSeek
$\Delta_1$ & -18.3 & -26.7 & -46.7 & -32.1\\
$\Delta_2$ & +0.7 & -6.7 & -13.3 & -17.9\\
AR & 84.5 & 55.6 & 71.4 & \textcolor{black!40}{n/a}\\
\addlinespace
\rowcolor{resultheader}\multicolumn{5}{l}{\textbf{Kimi K2.6}}\\
\MatchedSRKimi
$\Delta_1$ & -32.4 & -46.7 & -40.0 & -53.6\\
$\Delta_2$ & -1.4 & -20.0 & +13.3 & 0.0\\
AR & 68.9 & 25.0 & 100.0 & \textcolor{black!40}{n/a}\\
\addlinespace
\rowcolor{resultheader}\multicolumn{5}{l}{\textbf{qwen3.5-27b}}\\
\MatchedSRQwen
$\Delta_1$ & -27.5 & -46.7 & -33.3 & -53.6\\
$\Delta_2$ & -4.2 & 0.0 & 0.0 & +17.9\\
AR & 63.9 & 50.0 & 85.7 & \textcolor{black!40}{n/a}\\
\bottomrule
\end{tabularx}
\end{appendixpanels}
\end{table}

\begin{table}

\centering
\caption{Paired Offline and Online outcomes by task type. S/F means success offline and failure online. For Graft and Refresh, the AR denominator is S/S + S/F. Each model has 200 complete pairs.}
\label{tab:setting-pairs}
\small
\renewcommand{\arraystretch}{1.03}
\begin{appendixpanels}
\begin{tabular*}{\columnwidth}{@{\extracolsep{\fill}}lrrrr@{}}
\toprule
\textbf{Type} & S/S & S/F & F/S & F/F\\

\midrule
\rowcolor{resultheader}\multicolumn{5}{l}{\textbf{Terra}}\\
Graft & 122 & 4 & 4 & 12\\
Refresh: Replace & 12 & 1 & 1 & 1\\
Refresh: Retain & 9 & 0 & 3 & 3\\
Clarify & 20 & 0 & 3 & 5\\
\midrule
\rowcolor{resultheader}\multicolumn{5}{l}{\textbf{Gemini}}\\
Graft & 95 & 16 & 13 & 18\\
Refresh: Replace & 11 & 1 & 0 & 3\\
Refresh: Retain & 10 & 0 & 2 & 3\\
Clarify & 17 & 3 & 4 & 4\\
\midrule
\rowcolor{resultheader}\multicolumn{5}{l}{\textbf{Doubao}}\\
Graft & 104 & 12 & 7 & 19\\
Refresh: Replace & 7 & 6 & 1 & 1\\
Refresh: Retain & 10 & 0 & 1 & 4\\
Clarify & 11 & 3 & 9 & 5\\
\bottomrule\end{tabular*}
\appendixnextpanel
\begin{tabular*}{\columnwidth}{@{\extracolsep{\fill}}lrrrr@{}}
\toprule
\textbf{Type} & S/S & S/F & F/S & F/F\\

\midrule
\rowcolor{resultheader}\multicolumn{5}{l}{\textbf{DeepSeek}}\\
Graft & 82 & 15 & 16 & 29\\
Refresh: Replace & 5 & 4 & 3 & 3\\
Refresh: Retain & 5 & 2 & 0 & 8\\
Clarify & 13 & 6 & 1 & 8\\
\midrule
\rowcolor{resultheader}\multicolumn{5}{l}{\textbf{Kimi}}\\
Graft & 42 & 19 & 17 & 64\\
Refresh: Replace & 1 & 3 & 0 & 11\\
Refresh: Retain & 5 & 0 & 2 & 8\\
Clarify & 10 & 1 & 1 & 16\\
\midrule
\rowcolor{resultheader}\multicolumn{5}{l}{\textbf{qwen3.5-27b}}\\
Graft & 39 & 22 & 16 & 65\\
Refresh: Replace & 2 & 2 & 2 & 9\\
Refresh: Retain & 6 & 1 & 1 & 7\\
Clarify & 9 & 2 & 7 & 10\\
\bottomrule\end{tabular*}
\end{appendixpanels}
\end{table}

\begin{appendixcompacttable}

\centering\small
\caption{Setting-level results for qwen3.5-27b and UI-TARS. Entries give SR (\%) and successes out of the 200 scored runs in each setting.}
\label{tab:service-settings}
\renewcommand{\arraystretch}{1.12}
\begin{tabular*}{\linewidth}{@{\extracolsep{\fill}}lrrr@{}}
\toprule
Model & Computer-Use & Offline Ego Use & Online Ego Use\\
\midrule
qwen3.5-27b & 74.5 (149/200) & 41.5 (83/200) & 41.0 (82/200)\\
UI-TARS-1.5-7B & 0.0 (0/200) & 0.0 (0/200) & 0.0 (0/200)\\
\bottomrule\end{tabular*}
\end{appendixcompacttable}

\begin{table}

\centering\small
\caption{{Paired SR differences (\%) with 95\% recording-level bootstrap intervals. Each block specifies the subtraction order. $n$ counts complete task pairs from distinct recordings.}}
\label{tab:success-differences}
\renewcommand{\arraystretch}{1.12}
\setlength{\tabcolsep}{1.5pt}
\begin{appendixpanels}[.32\linewidth]
\begin{tabular*}{\columnwidth}{@{\extracolsep{\fill}}lrrr@{}}
\multicolumn{4}{@{}l}{Offline $-$ Computer-Use}\\[2pt]
\toprule
Model & $n$ & $\Delta$ & 95\% CI\\
\midrule
Doubao & 200 & $-15.5$ & $[-22.0, -8.5]$\\
Terra & 200 & $-10.0$ & $[-16.0, -4.0]$\\
Gemini & 200 & $-13.0$ & $[-19.0, -7.0]$\\
DeepSeek & 200 & $-23.0$ & $[-30.5, -16.0]$\\
Kimi & 200 & $-37.0$ & $[-45.0, -28.5]$\\
qwen3.5-27b & 200 & $-33.0$ & $[-41.5, -24.0]$\\
\bottomrule
\end{tabular*}
\appendixnextpanel
\begin{tabular*}{\columnwidth}{@{\extracolsep{\fill}}lrrr@{}}
\multicolumn{4}{@{}l}{Online $-$ Offline}\\[2pt]
\toprule
Model & $n$ & $\Delta$ & 95\% CI\\
\midrule
Doubao & 200 & $-1.5$ & $[-7.5, 4.5]$\\
Terra & 200 & $+3.0$ & $[-1.0, 7.0]$\\
Gemini & 200 & $-0.5$ & $[-7.0, 5.5]$\\
DeepSeek & 200 & $-3.5$ & $[-10.0, 3.0]$\\
Kimi & 200 & $-1.5$ & $[-8.5, 5.0]$\\
qwen3.5-27b & 200 & $-0.5$ & $[-7.5, 6.5]$\\
\bottomrule
\end{tabular*}
\appendixnextpanel
\begin{tabular*}{\columnwidth}{@{\extracolsep{\fill}}lrrr@{}}
\multicolumn{4}{@{}l}{Arrival: Wake-up $-$ Concurrent}\\[2pt]
\toprule
Model & $n$ & $\Delta$ & 95\% CI\\
\midrule
Terra & 98 & $-8.2$ & $[-15.3, -2.0]$\\
DeepSeek & 98 & $-20.4$ & $[-30.6, -10.2]$\\
Kimi & 98 & $-15.3$ & $[-26.5, -4.1]$\\
\bottomrule
\end{tabular*}
\end{appendixpanels}
\end{table}

\subsection{Online Input Conditions}
\label{app:online-input-conditions}
The comparisons below expand Table~\ref{tab:variables}. Arrival varies when the follow-up is delivered, Retention varies how older frames are represented, and Cue varies the information sent with a visual update.

\paragraph{Arrival.}
Table~\ref{tab:arrival-delivery} compares the same 98 Graft tasks when the follow-up arrives during execution or resumes a finished episode. Both conditions retain website state and share a 20-step budget.

\begin{appendixcompacttable}

\centering\small
\caption{Paired Arrival outcomes on 98 tasks per model. Columns indicate which conditions succeed.}
\label{tab:arrival-delivery}
\renewcommand{\arraystretch}{1.12}
\begin{tabular*}{\columnwidth}{@{\extracolsep{\fill}}lrrrr@{}}
\toprule
Model & Both & \shortstack{Concurrent\\only} & \shortstack{Wake-up\\only} & Neither\\
\midrule
Terra & 78 & 10 & 2 & 8\\
DeepSeek & 46 & 26 & 6 & 20\\
Kimi & 20 & 24 & 9 & 45\\
\bottomrule
\end{tabular*}
\end{appendixcompacttable}

{When Concurrent succeeds but Wake-up fails, agents often find a valid new answer and leave an obsolete selection in place. After checking all acceptable final outcomes, we find a complete new answer plus extra entries in 5 of 10 such pairs for Terra, 20 of 26 for DeepSeek, and 11 of 24 for Kimi. At least one extra entry was selected before the follow-up and still present at termination in 5, 19, and 11 pairs, respectively.}

{For example, under Wake-up, DeepSeek saves a T-shirt ironing lesson before the user asks for a tutorial on operating the iron. It saves a suitable tutorial at step 9 but ends at step 10 without removing the first lesson. Under Concurrent, it saves only the operation tutorial and succeeds.}

{Wake-up leaves fewer decision steps after the request update. Mean delivery occurs at step 2.49 under Concurrent and at steps 8.50, 8.09, and 7.18 under Wake-up for Terra, DeepSeek, and Kimi, respectively. In the example above, DeepSeek ends with steps still available.}

\paragraph{Retention.}
{Table~\ref{tab:retention-types} compares Full, Compressed, and Summarized visual histories on the same 47 tasks. Each task has a reviewed batch after the opening that leaves the recent-history window by the update or the step limit. In each model and condition, 97.9\% to 100.0\% of episodes reach the point where this older batch leaves the window. Compressed has the most successes for Terra and DeepSeek, while Summarized has the most for Kimi.}

\begin{appendixcompacttable}\centering

\caption{{Retention successes out of valid runs on 34 Graft, nine Refresh Replace, and four Refresh Retain tasks.}}
\label{tab:retention-types}\small
\renewcommand{\arraystretch}{1.12}
\begin{tabularx}{\columnwidth}{@{}Xrrr@{}}\toprule
History & Graft & Replace & Retain\\\midrule
\multicolumn{4}{l}{Terra}\\
Full & 30/34 & 6/9 & 4/4\\
Compressed & 30/34 & 8/9 & 4/4\\
Summarized & 28/34 & 8/9 & 4/4\\
\midrule
\multicolumn{4}{l}{DeepSeek}\\
Full & 22/34 & 3/9 & 2/4\\
Compressed & 23/34 & 5/9 & 2/4\\
Summarized & 18/34 & 3/9 & 2/4\\
\midrule
\multicolumn{4}{l}{Kimi}\\
Full & 18/34 & 1/9 & 1/4\\
Compressed & 14/34 & 1/9 & 2/4\\
Summarized & 19/34 & 2/9 & 2/4\\
\bottomrule\end{tabularx}\end{appendixcompacttable}

{The Summarized condition requires the agent to write descriptions while batches remain visible, then uses those descriptions when older images are removed. In the final model request, at least one older-batch description is unavailable in 0/47 Terra, 36/47 DeepSeek, and 41/47 Kimi episodes. These are missing memory entries within valid agent runs. The corresponding final requests contain 314 available and zero missing descriptions for Terra, 165 available and 121 missing for DeepSeek, and 83 available and 160 missing for Kimi. Performance with summarized history thus includes the agent's ability to construct the requested memory.}

{A summary can preserve an early misreading after the original frames have been removed. For example, DeepSeek's summary describes dark green leaves where the video shows peas being poured into a jug. With Summarized history it keeps searching for spinach and saves neither recipe. With Full history it switches to peas at step 9 and saves both requested recipes. In Kimi's bedding pair, both Full and Compressed find a valid duvet-cover lesson, but Compressed leaves the earlier pillowcase lesson in the playlist. Here, the error is in updating the saved list after the new target has been found.}

\paragraph{Cue.}
\label{app:cue-results}
{We compare three conditions on the same 30 Refresh tasks: Silent provides no alert, Cued adds an alert, and Cued + facts adds a verified scene description to the alert. Each condition includes 15 Replace and 15 Retain tasks. Table~\ref{tab:cue-delivery} shows that the alert alone changes overall success counts little. The paired results in Table~\ref{tab:cue-differences} show that adding scene descriptions can help in one task type and hurt in the other.}

All 135 Retain episodes have complete state evidence, and none holds a correct selection before the update. The sample therefore has no eligible observations for erroneous changes to an existing correct selection.

{DeepSeek's five gains from Cued to Cued + facts all occur on Replace tasks, while its three losses occur on Retain tasks. In the noodle-recipe pair, DeepSeek saves a vermicelli recipe with Cued. With Cued + facts, the description notes that the wearer adds yellow noodles to the vegetables. DeepSeek then searches for egg-noodle stir-fry and saves a valid recipe. In the shirt pair, DeepSeek keeps the original blue-and-white shirt as its target with Cued. With Cued + facts, the description says the wearer sets that shirt aside and lifts a red-and-black one. DeepSeek switches to the red-and-black shirt even though the visual change does not alter which shirt the user asked about.}

{The effect is not uniform across models. Terra moves in the other direction on the cleaning-cloth Retain task: Cued chooses the later orange cloth, whereas its Cued + facts run returns to a valid white-and-yellow cloth. Kimi gains two Retain tasks but loses one Replace task with Cued + facts. In addition, a cue cannot help a run that terminates before its delivery. Terra's tuna-recipe run with Cued + facts ends at step 5 before the step-12 update, while its Cued run continues to receive the update and succeeds. The paired SR contrast therefore includes both how delivered evidence is used and whether execution lasts long enough to receive it.}

\begin{appendixpanels}
\begin{table}

\centering\small
\caption{{Cue outcomes on 30 tasks per model and condition. Success counts final-state successes. Received counts episodes receiving the visual update.}}
\label{tab:cue-delivery}
\renewcommand{\arraystretch}{1.12}
\begin{tabularx}{\columnwidth}{@{}lXrr@{}}
\toprule
Model & Condition & Success & Received\\
\midrule
Terra & Silent & 25 & 30\\
Terra & Cued & 26 & 30\\
Terra & Cued + facts & 26 & 29\\
\midrule
DeepSeek & Silent & 13 & 27\\
DeepSeek & Cued & 15 & 28\\
DeepSeek & Cued + facts & 17 & 28\\
\midrule
Kimi & Silent & 8 & 26\\
Kimi & Cued & 8 & 27\\
Kimi & Cued + facts & 9 & 27\\
\bottomrule
\end{tabularx}
\end{table}
\appendixnextpanel
\begin{table}

\centering
\caption{{Paired Cue differences (\%) with 95\% recording-level bootstrap intervals. All rows use complete three-condition comparisons.}}
\label{tab:cue-differences}
\small
\renewcommand{\arraystretch}{1.12}
\setlength{\tabcolsep}{3pt}
\begin{tabularx}{\columnwidth}{@{}l*{2}{>{\centering\arraybackslash}X}@{}}
\toprule
 & Cued $-$ Silent & Cued + facts $-$ Cued\\
\midrule
\multicolumn{3}{l}{Terra}\\
Overall ($n=30$) & $+3.3\ [0.0,10.0]$ & $0.0\ [-13.3,13.3]$\\
Replace ($n=15$) & $+6.7\ [0.0,20.0]$ & $-13.3\ [-33.3,0.0]$\\
Retain ($n=15$) & $0.0\ [0.0,0.0]$ & $+13.3\ [0.0,33.3]$\\
\midrule
\multicolumn{3}{l}{DeepSeek}\\
Overall ($n=30$) & $+6.7\ [-10.0,23.3]$ & $+6.7\ [-13.3,23.4]$\\
Replace ($n=15$) & $-20.0\ [-40.0,0.0]$ & $+33.3\ [13.3,53.3]$\\
Retain ($n=15$) & $+33.3\ [13.3,60.0]$ & $-20.0\ [-40.0,0.0]$\\
\midrule
\multicolumn{3}{l}{Kimi}\\
Overall ($n=30$) & $0.0\ [-13.3,13.3]$ & $+3.3\ [-6.7,16.7]$\\
Replace ($n=15$) & $+13.3\ [0.0,33.3]$ & $-6.7\ [-20.0,0.0]$\\
Retain ($n=15$) & $-13.3\ [-33.3,0.0]$ & $+13.3\ [0.0,33.3]$\\
\bottomrule
\end{tabularx}
\end{table}
\end{appendixpanels}

\subsection{Clarification Behavior}
\label{app:clarification-behavior}
Table~\ref{tab:clarification-audit} examines whether agents had enough information to identify the requested item before selecting it. We review all 78 final-state successes among 112 scored Online Clarify runs from Terra, Gemini, Doubao, and DeepSeek on the same 28 tasks. For each run, we read the request, delivered frames, browser observations, and dialogue through the step at which the agent selected the item it kept. An undelivered reference reply is used only to check the intended referent after the fact.

We record \emph{Reply} if the agent receives a disambiguating answer before selection, and \emph{Evidence} if the observations identify the referent without a question or the same item satisfies all plausible interpretations. We record \emph{Unresolved} if at least two plausible referents still require different website choices. When the trace does not establish whether the available cues distinguish the referents, we record \emph{Unclear}.

\begin{appendixpanels}
\begin{table}
\centering
\caption{{Clarification outcomes among reviewed final-state successes. Each model has 28 scored Clarify runs.}}
\label{tab:clarification-audit}
\small
\renewcommand{\arraystretch}{1.12}
\setlength{\tabcolsep}{3pt}
\begin{tabular*}{\columnwidth}{@{\extracolsep{\fill}}lrrrrr@{}}
\toprule
Model & {Reviewed} & Reply & Evidence & Unresolved & Unclear\\
\midrule
Terra    & 23 & 1 & 1 & 20 & 1\\
Gemini   & 21 & 1 & 1 & 18 & 1\\
Doubao   & 20 & 1 & 1 & 17 & 1\\
DeepSeek & 14 & 0 & 0 & 14 & 0\\
\midrule
Total    & 78 & 3 & 3 & 69 & 3\\
\bottomrule
\end{tabular*}
\end{table}
\appendixnextpanel
\begin{table}

\centering
\caption{Clarify questioning in Offline and Online Ego Use. Each row covers 28 tasks. Success after reply gives successful runs out of those that receive the reviewed answer, or n/a when none does. No selection counts runs that never select a website item.}
\label{tab:setting-questions}
\small
\renewcommand{\arraystretch}{1.12}
\setlength{\tabcolsep}{2.5pt}
\begin{tabularx}{\columnwidth}{@{}Xrrrr@{}}
\toprule
\textbf{Model and setting} & Asked & \shortstack{Before\\selection} & \shortstack{No\\selection} & \shortstack{Success\\after reply}\\
\midrule
Terra: Offline & 2 & 2 & 0 & 1/2\\
Terra: Online & 1 & 1 & 0 & 1/1\\
Gemini: Offline & 2 & 2 & 0 & 2/2\\
Gemini: Online & 1 & 1 & 0 & 1/1\\
Doubao: Offline & 0 & 0 & 1 & n/a\\
Doubao: Online & 1 & 1 & 0 & 1/1\\
DeepSeek: Offline & 0 & 0 & 1 & n/a\\
DeepSeek: Online & 0 & 0 & 2 & n/a\\
Kimi: Offline & 0 & 0 & 1 & n/a\\
Kimi: Online & 0 & 0 & 0 & n/a\\
qwen3.5-27b: Offline & 0 & 0 & 0 & n/a\\
qwen3.5-27b: Online & 0 & 0 & 1 & n/a\\
\bottomrule
\end{tabularx}
\end{table}
\end{appendixpanels}

The reference remains unresolved at selection in 69 of the 78 reviewed successes. These trajectory labels leave the original final-state scores unchanged. Table~\ref{tab:setting-questions} compares when agents ask in Offline and Online Ego Use across all valid runs, including failures, on the same 28 Clarify tasks for the six main models. Before selection counts runs where the agent asks before first selecting a website item.

\subsection{Composite Changes}
\label{app:composite-comparison}
The 93-task comparison in Table~\ref{tab:composite} measures performance across combinations of Graft, Refresh, and Clarify. Figure~5 asks a different question: how one extra request update changes success on the same task. This section first describes the original tasks and their delivered information, then reports the paired results and the variant list.

Each visual or request update counts once, including updates affecting several attributes. Concurrent updates from different sources count separately. An update remains counted if it changes the acceptable choices at the time, even when a later update supersedes it. Reminders, unrelated scene cuts, and clarification replies are excluded. The set contains 26 tasks with one update, 47 with two, and 20 with three.

The 82 visual updates comprise 29 that invalidate an earlier choice, one that preserves it, and 52 that first establish a target. The 98 request updates comprise 71 refinements, eight additions, 13 replacements, and six withdrawals. Of 68 clarification opportunities, 40 concern position, ten sequence, nine variants, and nine category or purpose references. Table~\ref{tab:composite-types} groups tasks by the update types they contain. Each task is counted once per type and may occur in several rows.

\begin{table}
\centering
\caption{Composite task success by the types of information a task contains. Entries are successes out of valid runs. A task can appear in more than one row.}
\label{tab:composite-types}
\small
\renewcommand{\arraystretch}{1.12}
\begin{appendixpanels}
\begin{tabularx}{\columnwidth}{@{}Xrr@{}}
\toprule
\textbf{Type} & \textbf{Gemini} & \textbf{DeepSeek}\\
\midrule
Visual: invalidates a choice & 15/29 & 10/29\\
Visual: retains a choice & 1/1 & 1/1\\
Visual: first reveals a target & 33/38 & 16/38\\
\midrule
Request: refines & 51/62 & 27/62\\
Request: adds & 8/8 & 4/8\\
Request: replaces & 11/12 & 9/12\\
Request: withdraws & 4/6 & 3/6\\
\bottomrule
\end{tabularx}
\appendixnextpanel
\begin{tabularx}{\columnwidth}{@{}Xrr@{}}
\toprule
\textbf{Type} & \textbf{Gemini} & \textbf{DeepSeek}\\
\midrule
Clarification: position & 29/40 & 13/40\\
Clarification: sequence & 10/10 & 7/10\\
Clarification: variant & 7/9 & 0/9\\
Clarification: other & 7/9 & 4/9\\
\bottomrule
\end{tabularx}
\end{appendixpanels}
\end{table}

Gemini receives 80/82 visual updates and DeepSeek receives 79/82. Both receive all 98 request updates. Clarification replies are obtained on 26/68 and 9/68 tasks, respectively. Gemini reaches the step limit on 5/93 episodes and DeepSeek on 10/93. These delivery counts accompany final-state SR, which includes all valid episodes.

\paragraph{Original count groups.}
Figure~\ref{fig:composite-count-descriptive} groups the original 93 tasks by total update count. Those groups contain different tasks and differ in update type and construction batch.

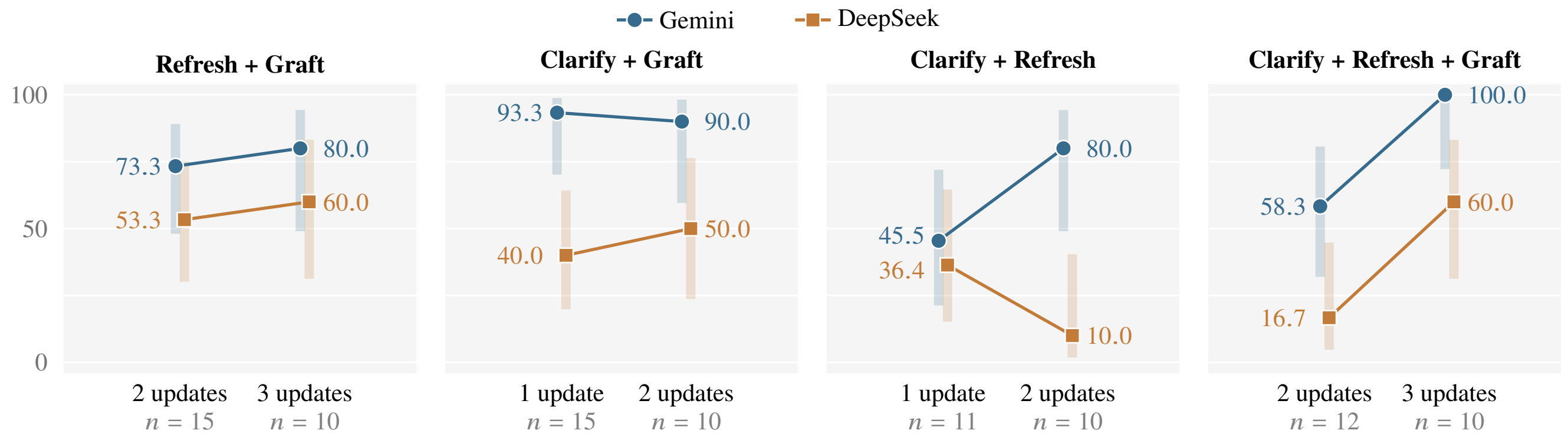
\begin{figure*}
\centering
\newcommand{\slopeframe}[5]{%
  \fill[black!4] (0,-4) rectangle (3.95,104);
  \foreach \t in {0,25,50,75,100} {\draw[white,line width=.6pt] (0,\t) -- (3.95,\t);}
  \node[anchor=south,font=\scriptsize\bfseries,inner sep=0pt] at (1.975,108) {#1};
  \node[anchor=north,align=center,inner sep=0pt] at (1.3,-8) {#2\\[-.5pt]{\color{black!50}$n=#3$}};
  \node[anchor=north,align=center,inner sep=0pt] at (2.7,-8) {#4\\[-.5pt]{\color{black!50}$n=#5$}};
}
\newcommand{\slopeci}[4]{\draw[#1,opacity=.2,line width=3pt] (#2,#3) -- (#2,#4);}
\newcommand{\slopemodel}[7]{%
  \draw[#1,line width=1pt] (#3,#4) -- (#3+1.4,#5);
  \node[#2] at (#3,#4) {};
  \node[#2] at (#3+1.4,#5) {};
  \node[anchor=east,inner sep=0pt,text=#1] at (1.1,#6) {\pgfmathprintnumber[fixed,precision=1,zerofill]{#4}};
  \node[anchor=west,inner sep=0pt,text=#1] at (2.9,#7) {\pgfmathprintnumber[fixed,precision=1,zerofill]{#5}};
}
\newcommand{\slopegem}[4]{\slopemodel{resultblue}{mgem}{1.25}{#1}{#2}{#3}{#4}}
\newcommand{\slopedeep}[4]{\slopemodel{resultorange}{mdeep}{1.35}{#1}{#2}{#3}{#4}}
\begin{tikzpicture}[x=1cm,y=.03cm,font=\scriptsize,
  mgem/.style={circle,fill=resultblue,draw=white,line width=.5pt,inner sep=0pt,minimum size=5pt},
  mdeep/.style={rectangle,fill=resultorange,draw=white,line width=.5pt,inner sep=0pt,minimum size=4.6pt}]
  \draw[resultblue,line width=1pt] (6.75,128) -- (7.15,128);
  \node[mgem] at (6.95,128) {};
  \node[anchor=west,inner sep=1.5pt] at (7.17,128) {Gemini};
  \draw[resultorange,line width=1pt] (8.75,128) -- (9.15,128);
  \node[mdeep] at (8.95,128) {};
  \node[anchor=west,inner sep=1.5pt] at (9.17,128) {DeepSeek};
  \foreach \t in {0,50,100} {\node[anchor=east,text=black!55] at (.5,\t) {\t};}
  \begin{scope}[xshift=.55cm]
    \slopeframe{Refresh + Graft}{2 updates}{15}{3 updates}{10}
    \slopeci{resultblue}{1.25}{48.05}{89.1}
    \slopeci{resultblue}{2.65}{49.02}{94.33}
    \slopeci{resultorange}{1.35}{30.12}{75.19}
    \slopeci{resultorange}{2.75}{31.27}{83.18}
    \slopedeep{53.33}{60.0}{53.33}{60.0}
    \slopegem{73.33}{80.0}{73.33}{80.0}
  \end{scope}
  \begin{scope}[xshift=4.83cm]
    \slopeframe{Clarify + Graft}{1 update}{15}{2 updates}{10}
    \slopeci{resultblue}{1.25}{70.18}{98.81}
    \slopeci{resultblue}{2.65}{59.58}{98.21}
    \slopeci{resultorange}{1.35}{19.82}{64.25}
    \slopeci{resultorange}{2.75}{23.66}{76.34}
    \slopedeep{40.0}{50.0}{40.0}{50.0}
    \slopegem{93.33}{90.0}{93.33}{90.0}
  \end{scope}
  \begin{scope}[xshift=9.11cm]
    \slopeframe{Clarify + Refresh}{1 update}{11}{2 updates}{10}
    \slopeci{resultblue}{1.25}{21.27}{71.99}
    \slopeci{resultblue}{2.65}{49.02}{94.33}
    \slopeci{resultorange}{1.35}{15.17}{64.62}
    \slopeci{resultorange}{2.75}{1.79}{40.42}
    \slopedeep{36.36}{10.0}{34.5}{10.0}
    \slopegem{45.45}{80.0}{47.5}{80.0}
  \end{scope}
  \begin{scope}[xshift=13.39cm]
    \slopeframe{Clarify + Refresh + Graft}{2 updates}{12}{3 updates}{10}
    \slopeci{resultblue}{1.25}{31.95}{80.67}
    \slopeci{resultblue}{2.65}{72.25}{100.0}
    \slopeci{resultorange}{1.35}{4.7}{44.8}
    \slopeci{resultorange}{2.75}{31.27}{83.18}
    \slopedeep{16.67}{60.0}{16.67}{60.0}
    \slopegem{58.33}{100.0}{58.33}{100.0}
  \end{scope}
\end{tikzpicture}
\caption{Descriptive success rate (\%) by update count in the original 93 tasks. Different count groups contain different tasks and are confounded by update type and construction batch. These slopes do not estimate a within-task count effect. $n$ counts tasks from distinct recordings. Lines connect each model's success rates within a combination, and shaded bars are 95\% Wilson intervals. Clarification does not count as an update.}
\label{fig:composite-count-descriptive}
\end{figure*}

\paragraph{Within-task request-update count control.}
We screen 49 original tasks before running any variant and retain 46 pairs: 17 Clarify + Graft, 13 Refresh + Graft, and 16 three-source tasks. These 46 variants are additional controlled task versions, separate from the 293 benchmark tasks and their matched counterparts. The original 93 tasks and their 186 valid episodes remain unchanged. Each new task references its original and modifies only user-request events. We inherit the original recording window, opening request, visual changes and times, clarification, frozen website, accepted final states, and 30-step budget. Clarification remains question-triggered, so its actual receipt can differ across runs.

In 25 merge pairs, two requests become one natural request delivered at the original second request time. In 21 insertion pairs, one intermediate request precedes the original request without changing its time or text. The intermediate request must exclude at least one complete website state that was acceptable at that time. We check this against public facts in the original frozen website, including any earlier visual change, and retain a valid state as a witness. The insertions comprise 16 progressive tightenings, one split addition (CG-03), and four quantity revisions (CG-04, CG-06, CG-17, CRG-01). Every new event occupies a distinct eight-second replay batch. The final scheduled information time is unchanged. Request-to-request gaps have a median of 24 seconds in merge originals (range 8 to 77) and 20 seconds in insertion variants (range 8 to 80). Five insertion pairs retain adjacent replay batches because the opening or visual change leaves no earlier slot. Delivery statistics use the actual trajectories, including standby jumps and question-triggered replies.

Three candidates are excluded before evaluation. In CG-12, the proposed duration limit makes the original request redundant. In RG-02, we could not verify an acceptable choice that the proposed request rules out before the visual change. RG-18 has no free replay batch. Clarify + Refresh keeps the request fixed while new information comes from visual changes and clarification replies, so it does not enter the request-update count control. All four quantity revisions pass the naturalness review. Tables~\ref{tab:count-merge-list} and~\ref{tab:count-insert-list} list the frozen variants.

We evaluate both models on the same 92 frozen task versions, with three runs per version and 276 valid runs per model. The first original-task runs are reused from the formal evaluation. Gemini's reviewed development episodes supply its first variant runs. SR includes valid unsuccessful episodes and excludes invalid service attempts.

Across the three rounds, both models receive all 201 variant request updates. Gemini receives 83/87 variant visual updates and DeepSeek receives 82/87. Four visual batches per model are skipped on standby, and one further DeepSeek batch is not delivered before execution ends. Clarification replies are obtained in 38/99 and 15/99 opportunities, respectively.

Table~\ref{tab:count-paired-results} gives the success rates behind the differences plotted in Figure~5 of the main paper. In merge pairs, the original task has more request updates. In insertion pairs, the variant has more. We report the two comparisons separately because they change different parts of the request sequence.

We average the three runs for each task version before resampling source recordings as described in Section~\ref{app:experiment-details}. Each combination and direction contains six to ten independent pairs. Table~\ref{tab:count-stability} reports the difference in each round and counts task versions with mixed outcomes across runs.

\begin{table*}
\centering\small
\caption{Three-run paired request-update count results. Each row uses the same tasks in both conditions. SR averages three runs per task version. $n$ counts independent task pairs. SR and $\Delta$ are reported in \%. $\Delta$ is more minus fewer. Intervals use paired recording-level bootstrap. More-only and fewer-only count discordant task-pair repetitions, with $3n$ paired repetitions per row.}
\label{tab:count-paired-results}
\renewcommand{\arraystretch}{1.12}
\setlength{\tabcolsep}{5pt}
\begin{tabular*}{\textwidth}{@{\extracolsep{\fill}}lllrrrrrr@{}}
\toprule
Direction & Combination & Model & $n$ & Fewer SR & More SR & $\Delta$ [95\% CI] & More-only & Fewer-only\\
\midrule
Merge & C+G & Gemini & 10 & 66.7 & 76.7 & $ 10.0\ [-6.7,26.7] $ & 4 & 1\\
Merge & C+G & DeepSeek & 10 & 56.7 & 50.0 & $ -6.7\ [-30.0,16.7] $ & 3 & 5\\
Merge & R+G & Gemini & 6 & 72.2 & 77.8 & $ 5.6\ [0.0,16.7] $ & 1 & 0\\
Merge & R+G & DeepSeek & 6 & 55.6 & 72.2 & $ 16.7\ [0.0,38.9] $ & 3 & 0\\
Merge & C+R+G & Gemini & 9 & 77.8 & 85.2 & $ 7.4\ [0.0,14.8] $ & 2 & 0\\
Merge & C+R+G & DeepSeek & 9 & 55.6 & 44.4 & $ -11.1\ [-22.2,-3.7] $ & 2 & 5\\
\midrule
Insert & C+G & Gemini & 7 & 85.7 & 85.7 & $ 0.0\ [-14.3,14.3] $ & 1 & 1\\
Insert & C+G & DeepSeek & 7 & 38.1 & 42.9 & $ 4.8\ [-28.6,38.1] $ & 4 & 3\\
Insert & R+G & Gemini & 7 & 95.2 & 100.0 & $ 4.8\ [0.0,14.3] $ & 1 & 0\\
Insert & R+G & DeepSeek & 7 & 52.4 & 61.9 & $ 9.5\ [0.0,19.0] $ & 2 & 0\\
Insert & C+R+G & Gemini & 7 & 57.1 & 57.1 & $ 0.0\ [-19.0,19.0] $ & 3 & 3\\
Insert & C+R+G & DeepSeek & 7 & 28.6 & 33.3 & $ 4.8\ [0.0,14.3] $ & 1 & 0\\
\bottomrule
\end{tabular*}
\end{table*}

Table~\ref{tab:count-stability} shows that the same request contrast can yield different SR differences across runs. For three-source insertion, Gemini's more-minus-fewer difference is $-14.3\%$, $28.6\%$, and $-14.3\%$ in the three rounds. For Clarify + Graft insertion, DeepSeek's difference is $28.6\%$, $14.3\%$, and $-28.6\%$. Across the three runs, 17/92 Gemini task versions and 24/92 DeepSeek task versions have both successful and unsuccessful outcomes.

\begin{table*}
\centering\small
\caption{Request-update count variability across three rounds. Each $\Delta$ (\%) is more minus fewer SR on the same $n$ independent pairs. SD is the sample standard deviation of the three round differences. Mixed outcomes count task versions with both success and failure across the three runs, separately for the fewer and more sides.}
\label{tab:count-stability}
\renewcommand{\arraystretch}{1.12}
\setlength{\tabcolsep}{5pt}
\begin{tabular*}{\textwidth}{@{\extracolsep{\fill}}lllrrrrrrr@{}}
\toprule
Direction & Combination & Model & $n$ & $\Delta_0$ & $\Delta_1$ & $\Delta_2$ & SD & \multicolumn{2}{c}{Mixed outcomes}\\
\cmidrule(lr){9-10}
& & & & & & & & Fewer & More\\
\midrule
Merge & C+G & Gemini & 10 & 20.0 & 10.0 & 0.0 & 10.0 & 2 & 3\\
Merge & C+G & DeepSeek & 10 & -10.0 & 0.0 & -10.0 & 5.8 & 3 & 4\\
Merge & R+G & Gemini & 6 & 16.7 & 0.0 & 0.0 & 9.6 & 1 & 1\\
Merge & R+G & DeepSeek & 6 & 16.7 & 0.0 & 33.3 & 16.7 & 1 & 1\\
Merge & C+R+G & Gemini & 9 & 22.2 & 0.0 & 0.0 & 12.8 & 0 & 2\\
Merge & C+R+G & DeepSeek & 9 & 11.1 & -33.3 & -11.1 & 22.2 & 4 & 2\\
\midrule
Insert & C+G & Gemini & 7 & 0.0 & 14.3 & -14.3 & 14.3 & 0 & 2\\
Insert & C+G & DeepSeek & 7 & 28.6 & 14.3 & -28.6 & 29.7 & 2 & 4\\
Insert & R+G & Gemini & 7 & 0.0 & 0.0 & 14.3 & 8.2 & 1 & 0\\
Insert & R+G & DeepSeek & 7 & 14.3 & 14.3 & 0.0 & 8.2 & 1 & 1\\
Insert & C+R+G & Gemini & 7 & -14.3 & 28.6 & -14.3 & 24.7 & 3 & 2\\
Insert & C+R+G & DeepSeek & 7 & 14.3 & 0.0 & 0.0 & 8.2 & 0 & 1\\
\bottomrule
\end{tabular*}
\end{table*}

\begin{appendixpanels}
\begin{table}
\centering\small
\caption{Merged-request variants. One request replaces two at the original second request time. Times are in source-video seconds. Counts include visual and request updates.}
\label{tab:count-merge-list}
\renewcommand{\arraystretch}{1.04}
\setlength{\tabcolsep}{3pt}
\begin{tabular*}{\columnwidth}{@{\extracolsep{\fill}}llrr@{}}
\toprule
Pair & Edit & Time (s) & Fewer/more\\
\midrule
CG-10 & Merge & 24.5 & 1/2\\
CG-16 & Merge & 32 & 1/2\\
CG-19 & Merge & 176.7 & 1/2\\
CG-20 & Merge & 772 & 1/2\\
CG-21 & Merge & 574 & 1/2\\
CG-22 & Merge & 144 & 1/2\\
CG-23 & Merge & 136 & 1/2\\
CG-24 & Merge & 65 & 1/2\\
CG-25 & Merge & 509 & 1/2\\
CG-26 & Merge & 2037 & 1/2\\
CRG-13 & Merge & 406 & 2/3\\
CRG-14 & Merge & 124 & 2/3\\
CRG-15 & Merge & 104 & 2/3\\
CRG-16 & Merge & 1456 & 2/3\\
CRG-17 & Merge & 570 & 2/3\\
CRG-19 & Merge & 552 & 2/3\\
CRG-20 & Merge & 938 & 2/3\\
CRG-21 & Merge & 492 & 2/3\\
CRG-22 & Merge & 136 & 2/3\\
RG-06 & Merge & 451.2 & 2/3\\
RG-23 & Merge & 43 & 2/3\\
RG-24 & Merge & 165 & 2/3\\
RG-25 & Merge & 104 & 2/3\\
RG-26 & Merge & 66 & 2/3\\
RG-28 & Merge & 97 & 2/3\\
\bottomrule
\end{tabular*}
\end{table}
\appendixnextpanel
\begin{table}
\centering\small
\caption{Inserted-request variants. The new intermediate request precedes the unchanged original request. Times are in source-video seconds. Counts include visual and request updates.}
\label{tab:count-insert-list}
\renewcommand{\arraystretch}{1.04}
\setlength{\tabcolsep}{3pt}
\begin{tabular*}{\columnwidth}{@{\extracolsep{\fill}}llrr@{}}
\toprule
Pair & Intermediate type & Time (s) & Fewer/more\\
\midrule
CG-01 & Tighten & 76 & 1/2\\
CG-03 & Split addition & 41 & 1/2\\
CG-04 & Quantity revision & 16 & 1/2\\
CG-06 & Quantity revision & 59 & 1/2\\
CG-08 & Tighten & 28 & 1/2\\
CG-11 & Tighten & 42.5 & 1/2\\
CG-17 & Quantity revision & 120 & 1/2\\
CRG-01 & Quantity revision & 693 & 2/3\\
CRG-02 & Tighten & 789 & 2/3\\
CRG-03 & Tighten & 392.2 & 2/3\\
CRG-08 & Tighten & 293.2 & 2/3\\
CRG-09 & Tighten & 696.2 & 2/3\\
CRG-10 & Tighten & 35.2 & 2/3\\
CRG-12 & Tighten & 88.2 & 2/3\\
RG-04 & Tighten & 238.2 & 2/3\\
RG-05 & Tighten & 482 & 2/3\\
RG-07 & Tighten & 913 & 2/3\\
RG-08 & Tighten & 85 & 2/3\\
RG-10 & Tighten & 21 & 2/3\\
RG-11 & Tighten & 128 & 2/3\\
RG-12 & Tighten & 109 & 2/3\\
\bottomrule
\end{tabular*}
\end{table}
\end{appendixpanels}

\paragraph{Paired trajectories.}
An extra request can prompt the agent to revise an earlier choice. For example, in Gemini's third RG-11 run, the inserted five-minute limit arrives at step 9. Gemini removes a 6:48 rules video and saves a valid 1:41 explanation at step 14. With fewer requests, it opens the correct short video but stops without saving it or removing the earlier selection.

An extra request can also be applied to the wrong target. For example, in DeepSeek's second CRG-17 run, the agent switches from mixing bowls to kettles after the glass-body requirement arrives, without asking which object the user means. It finishes with a kettle and the correct fruit purée. With the requests merged, it keeps a valid glass mixing bowl and the fruit purée.

\subsection{Replay Pace}
\label{app:protocol-sensitivity}
The pace comparison uses 120 tasks: 69 Graft, 23 Refresh, and 28 Clarify. Each step delivers eight frames spanning 4, 8, or 16 seconds, corresponding to sampling rates of 2, 1, and 0.5\,fps. The decision budget is 20 steps. Tasks whose change enters the opening batch at any tested pace are excluded. Each model has 360 scored episodes.

Table~\ref{tab:pace-results} reports success and update receipt. Each step contains the same number of frames, so advancing more source seconds spaces the frames farther apart and covers more of the recording within the decision budget. At 16 seconds per step, all 92 Graft and Refresh updates reach each model. At 4 seconds per step, some runs never receive the update. More coverage does not yield the same best setting for every model: Terra has the most successes at 8 seconds, while DeepSeek and Kimi have the most at 16 seconds.

\begin{appendixcompacttable}

\centering\small
\caption{{Pace sensitivity on 120 tasks per model. Receipt counts Graft and Refresh episodes receiving their update out of 92. Delivery step is the median among received updates. $\Delta$ (\%) is SR minus the SR at 8 seconds per step, with a recording-level bootstrap 95\% interval.}}
\label{tab:pace-results}
\renewcommand{\arraystretch}{1.12}
\setlength{\tabcolsep}{2pt}
\begin{tabular*}{\columnwidth}{@{\extracolsep{\fill}}llrrrr@{}}
\toprule
Model & Pace & Success & Receipt & Step & $\Delta$ [95\% CI]\\
\midrule
Terra & 4 s & 98/120 & 85/92 & 6 & $-7.5\ [-13.3,-1.7]$\\
Terra & 8 s & 107/120 & 92/92 & 5 & Reference\\
Terra & 16 s & 106/120 & 92/92 & 3 & $-0.8\ [-5.0,3.3]$\\
DeepSeek & 4 s & 60/120 & 84/92 & 7 & $-7.5\ [-16.7,2.5]$\\
DeepSeek & 8 s & 69/120 & 89/92 & 4 & Reference\\
DeepSeek & 16 s & 74/120 & 92/92 & 3 & $+4.2\ [-3.3,10.8]$\\
Kimi & 4 s & 49/120 & 75/92 & 6 & $0.0\ [-8.3,9.2]$\\
Kimi & 8 s & 49/120 & 88/92 & 5 & Reference\\
Kimi & 16 s & 58/120 & 92/92 & 3 & $+7.5\ [0.0,15.8]$\\
\bottomrule
\end{tabular*}
\end{appendixcompacttable}

\section{Execution Cost}
\label{app:execution-cost}
\subsection{Resource Accounting}
Costs are measured on the same episodes used for the success results. Input and output tokens are summed over logged requests, including repeated history. Per-task total token cost is their sum, averaged over all valid task episodes, including failures. Input totals include cached tokens and any image tokens counted by the provider. Output totals use the provider-reported completion count. Images are counted each time they are sent. Token comparisons are made within a model because token accounting differs across providers.

Each decision step in these episodes corresponds to one logged model request. Transport retries within a request are excluded from this count. Browser-operation counts include unsuccessful attempts but exclude questions, completion declarations, standby, and control waits. Wall-clock duration includes model-service waits, browser execution, and runtime overhead. It is distinct from source-video replay time and is not an isolated measure of inference latency. Serving routes and runtime load were not controlled for speed comparisons.

Table~\ref{tab:cost-overview} separates the input and output tokens that make up the totals in Figure~4b of the main paper. The main-task rows cover the same 200 episodes per model and setting, including valid failures. The table also gives resource use for the 93 original composite tasks and the three-run request-update comparison. Table~\ref{tab:controlled-costs} reports cost for the online input conditions in Table~\ref{tab:variables}. All listed episodes have complete token and duration records.

\begin{appendixlongtable}{llrrrrrrr}
\caption{{Resource use per valid episode. Input, output, and total tokens are in thousands. Total sums input and output per task. Time columns give mean and median wall-clock seconds. Each step corresponds to one logged model request. Counts refer to each analysis group. A run can appear in more than one group.}}
\label{tab:cost-overview}\\
\toprule
Model & Setting & $n$ & Steps & Input (k) & Output (k) & Total (k) & Mean time & Median time\\
\midrule
\endfirsthead
\appendixcontinued{9}
\toprule
Model & Setting & $n$ & Steps & Input (k) & Output (k) & Total (k) & Mean time & Median time\\
\midrule
\endhead
\midrule
\endfoot
\bottomrule
\endlastfoot
\rowcolor{resultheader}\multicolumn{9}{l}{\textbf{Main tasks, 20-step budget}}\\*
Terra & Computer-Use & 200 & 7.66 & 126.03 & 0.263 & 126.29 & 35.9 & 28.1\\*
Terra & Offline & 200 & 9.46 & 432.50 & 0.485 & 432.99 & 105.0 & 83.9\\*
Terra & Online & 200 & 9.84 & 298.04 & 0.513 & 298.55 & 80.2 & 67.3\\
\addlinespace
DeepSeek & Computer-Use & 200 & 9.42 & 175.11 & 3.064 & 178.17 & 34.3 & 21.8\\*
DeepSeek & Offline & 200 & 10.55 & 341.88 & 5.470 & 347.35 & 72.5 & 52.0\\*
DeepSeek & Online & 200 & 11.01 & 284.16 & 6.073 & 290.23 & 68.7 & 48.8\\
\addlinespace
Kimi & Computer-Use & 200 & 6.44 & 84.02 & 0.160 & 84.18 & 42.3 & 33.3\\*
Kimi & Offline & 200 & 7.75 & 328.09 & 0.188 & 328.28 & 279.9 & 198.8\\*
Kimi & Online & 200 & 8.44 & 226.64 & 0.201 & 226.84 & 159.5 & 110.2\\
\addlinespace
qwen3.5-27b & Computer-Use & 200 & 8.13 & 132.47 & 0.140 & 132.61 & 62.4 & 41.5\\*
qwen3.5-27b & Offline & 200 & 10.15 & 425.37 & 0.156 & 425.52 & 274.5 & 169.6\\*
qwen3.5-27b & Online & 200 & 9.44 & 288.44 & 0.155 & 288.59 & 149.2 & 84.9\\
\addlinespace
Gemini & Computer-Use & 200 & 9.01 & 177.89 & 1.785 & 179.68 & 65.7 & 50.4\\*
Gemini & Offline & 200 & 10.43 & 502.60 & 9.629 & 512.23 & 229.1 & 151.6\\*
Gemini & Online & 200 & 10.00 & 310.90 & 6.700 & 317.60 & 161.4 & 107.2\\
\addlinespace
Doubao & Computer-Use & 200 & 6.93 & 104.65 & 0.615 & 105.26 & 78.0 & 59.5\\*
Doubao & Offline & 200 & 7.72 & 392.46 & 1.862 & 394.32 & 211.9 & 164.1\\*
Doubao & Online & 200 & 8.40 & 248.39 & 1.660 & 250.05 & 123.7 & 65.3\\
\addlinespace
UI-TARS & Computer-Use & 200 & 20.00 & 262.70 & 2.035 & 264.73 & 78.3 & 46.3\\*
UI-TARS & Offline & 200 & 20.00 & 698.92 & 1.205 & 700.12 & 232.6 & 197.9\\*
UI-TARS & Online & 200 & 20.00 & 485.16 & 1.307 & 486.47 & 154.9 & 116.2\\
\addlinespace
\rowcolor{resultheader}\multicolumn{9}{l}{\textbf{Composite tasks, 30-step budget}}\\*
Gemini & Online & 93 & 14.71 & 589.46 & 12.225 & 601.69 & 222.0 & 157.9\\*
DeepSeek & Online & 93 & 17.06 & 710.30 & 21.211 & 731.52 & 189.2 & 119.8\\
\addlinespace
\rowcolor{resultheader}\multicolumn{9}{l}{\textbf{Request-update count comparison, three runs per version}}\\*
Gemini & Fewer request updates & 138 & 13.79 & 573.20 & 9.280 & 582.48 & 217.6 & 139.4\\*
Gemini & More request updates & 138 & 13.36 & 531.96 & 8.756 & 540.71 & 201.8 & 135.6\\*
DeepSeek & Fewer request updates & 138 & 16.57 & 689.13 & 14.121 & 703.25 & 168.7 & 101.9\\*
DeepSeek & More request updates & 138 & 16.59 & 718.77 & 14.566 & 733.34 & 163.3 & 87.7\\
\end{appendixlongtable}

{Under the controlled history comparison, Summarized uses fewer input tokens than Full for all three models. DeepSeek nevertheless takes longer with Summarized.}

\begin{appendixlongtable}{llrrrrrr}
\caption{{Resource use under controlled online conditions. Each task-condition pair has one valid episode with a 20-step budget. Input and output tokens are in thousands. Time columns give mean and median wall-clock seconds. Each model uses the same tasks within a comparison.}}
\label{tab:controlled-costs}\\
\toprule
Comparison & Condition & $n$ & Steps & Input (k) & Output (k) & Mean (s) & Median (s)\\
\midrule
\endfirsthead
\appendixcontinued{8}
\toprule
Comparison & Condition & $n$ & Steps & Input (k) & Output (k) & Mean (s) & Median (s)\\
\midrule
\endhead
\midrule
\endfoot
\bottomrule
\endlastfoot
\rowcolor{resultheader}\multicolumn{8}{l}{\textbf{DeepSeek}}\\*
Arrival & Concurrent & 98 & 10.29 & 230.94 & 4.654 & 57.1 & 43.0\\*
 & Wake-up & 98 & 13.83 & 348.71 & 5.162 & 165.6 & 85.0\\*
\addlinespace
Retention & Full & 47 & 12.60 & 382.38 & 5.866 & 135.5 & 72.2\\*
 & Compressed & 47 & 12.64 & 351.47 & 5.542 & 71.2 & 59.5\\*
 & Summarized & 47 & 11.94 & 294.96 & 4.189 & 234.4 & 58.9\\*
\addlinespace
Cue & Silent & 30 & 9.53 & 246.63 & 4.580 & 58.5 & 44.5\\*
 & Cued & 30 & 8.40 & 173.04 & 4.032 & 282.7 & 166.7\\*
 & Cued + facts & 30 & 8.70 & 192.33 & 5.409 & 260.4 & 119.0\\*
\addlinespace
Pace & 4 s & 120 & 12.53 & 345.11 & 5.371 & 88.7 & 58.3\\*
 & 8 s & 120 & 11.62 & 312.82 & 5.978 & 70.0 & 55.2\\*
 & 16 s & 120 & 10.75 & 293.77 & 7.070 & 92.0 & 48.5\\
\addlinespace
\rowcolor{resultheader}\multicolumn{8}{l}{\textbf{Kimi}}\\*
Arrival & Concurrent & 98 & 8.20 & 200.50 & 0.198 & 164.2 & 87.2\\*
 & Wake-up & 98 & 12.06 & 309.52 & 0.307 & 201.7 & 176.6\\*
\addlinespace
Retention & Full & 47 & 9.30 & 258.56 & 0.247 & 190.5 & 183.9\\*
 & Compressed & 47 & 9.49 & 256.88 & 0.233 & 155.2 & 155.0\\*
 & Summarized & 47 & 9.09 & 196.05 & 0.402 & 145.1 & 120.6\\*
\addlinespace
Cue & Silent & 30 & 6.33 & 179.03 & 0.148 & 123.3 & 104.5\\*
 & Cued & 30 & 7.20 & 251.25 & 0.150 & 169.9 & 98.8\\*
 & Cued + facts & 30 & 6.43 & 172.12 & 0.155 & 153.2 & 113.8\\*
\addlinespace
Pace & 4 s & 120 & 9.03 & 254.12 & 0.222 & 195.6 & 144.6\\*
 & 8 s & 120 & 8.69 & 237.23 & 0.202 & 148.6 & 137.7\\*
 & 16 s & 120 & 8.37 & 202.99 & 0.185 & 143.7 & 114.0\\
\addlinespace
\rowcolor{resultheader}\multicolumn{8}{l}{\textbf{Terra}}\\*
Arrival & Concurrent & 98 & 9.55 & 255.36 & 0.431 & 75.4 & 58.9\\*
 & Wake-up & 98 & 12.89 & 388.26 & 0.563 & 96.9 & 86.5\\*
\addlinespace
Retention & Full & 47 & 11.70 & 471.87 & 0.550 & 110.2 & 90.1\\*
 & Compressed & 47 & 11.68 & 390.60 & 0.556 & 93.6 & 80.4\\*
 & Summarized & 47 & 12.13 & 393.24 & 1.110 & 86.5 & 69.3\\*
\addlinespace
Cue & Silent & 30 & 9.93 & 369.14 & 0.641 & 89.3 & 69.6\\*
 & Cued & 30 & 9.50 & 346.11 & 0.611 & 85.3 & 74.1\\*
 & Cued + facts & 30 & 9.13 & 338.69 & 0.518 & 84.6 & 59.8\\*
\addlinespace
Pace & 4 s & 120 & 10.35 & 343.94 & 0.543 & 90.0 & 72.7\\*
 & 8 s & 120 & 9.99 & 318.09 & 0.497 & 81.4 & 67.8\\*
 & 16 s & 120 & 9.29 & 271.65 & 0.478 & 72.6 & 56.4\\*
\addlinespace
\end{appendixlongtable}

\subsection{Paired Comparison}
Table~\ref{tab:setting-costs} compares Offline and Online resource use across all 200 valid task pairs per model, including failures. Online uses fewer input tokens and first-person images for all six models, whereas mean output token use changes in both directions across models. The lower total token use in Figure~4b of the main paper therefore reflects lower input cost rather than consistently shorter responses.

\begingroup\setlength{\tabcolsep}{0pt}
\begin{appendixlongtable}{p{.488\linewidth}p{.488\linewidth}}
\caption{Mean resource use on all valid Offline and Online task pairs. Differences are Online minus Offline with recording-level bootstrap 95\% intervals. Each model has 200 pairs.}
\label{tab:setting-costs}\\
\endfirsthead
\appendixcontinued{2}
\endhead
\begin{appendixcostpanel}
\multicolumn{5}{l}{\textbf{{Terra}}}\\
Browser operations & 8.46 & 8.57 & $+0.12$ & $[-0.30, 0.53]$\\
Decision steps & 9.46 & 9.84 & $+0.37$ & $[-0.07, 0.79]$\\
Input tokens (k) & 432.50 & 298.04 & $-134.47$ & $[-163.77, -104.32]$\\
Output tokens (k) & 0.485 & 0.513 & $+0.028$ & $[-0.023, 0.076]$\\
First-person images & 205.22 & 86.53 & $-118.69$ & $[-135.92, -103.23]$\\
Website screenshots & 9.46 & 9.84 & $+0.37$ & $[-0.07, 0.79]$\\
Wall-clock time (s) & 105.04 & 80.23 & $-24.81$ & $[-32.50, -17.52]$\\
\end{appendixcostpanel}&
\begin{appendixcostpanel}
\multicolumn{5}{l}{\textbf{{DeepSeek}}}\\
Browser operations & 7.64 & 7.93 & $+0.29$ & $[-0.23, 0.82]$\\
Decision steps & 10.55 & 11.01 & $+0.46$ & $[-0.16, 1.10]$\\
Input tokens (k) & 341.88 & 284.16 & $-57.71$ & $[-90.39, -25.20]$\\
Output tokens (k) & 5.470 & 6.073 & $+0.604$ & $[-0.950, 2.238]$\\
First-person images & 235.60 & 103.42 & $-132.19$ & $[-161.44, -109.18]$\\
Website screenshots & 10.55 & 11.01 & $+0.46$ & $[-0.16, 1.10]$\\
Wall-clock time (s) & 72.50 & 68.72 & $-3.78$ & $[-14.06, 6.68]$\\
\end{appendixcostpanel}\\[3pt]
\begin{appendixcostpanel}
\multicolumn{5}{l}{\textbf{{Doubao}}}\\
Browser operations & 6.74 & 7.21 & $+0.47$ & $[-0.06, 0.98]$\\
Decision steps & 7.72 & 8.40 & $+0.69$ & $[0.17, 1.19]$\\
Input tokens (k) & 392.46 & 248.39 & $-144.06$ & $[-190.84, -100.60]$\\
Output tokens (k) & 1.862 & 1.660 & $-0.202$ & $[-0.615, 0.225]$\\
First-person images & 176.69 & 70.46 & $-106.23$ & $[-128.88, -86.60]$\\
Website screenshots & 7.72 & 8.40 & $+0.69$ & $[0.17, 1.19]$\\
Wall-clock time (s) & 211.94 & 123.72 & $-88.22$ & $[-111.42, -64.91]$\\
\end{appendixcostpanel}&
\begin{appendixcostpanel}
\multicolumn{5}{l}{\textbf{{Kimi}}}\\
Browser operations & 6.71 & 7.19 & $+0.47$ & $[0.04, 0.90]$\\
Decision steps & 7.75 & 8.44 & $+0.69$ & $[0.26, 1.13]$\\
Input tokens (k) & 328.09 & 226.64 & $-101.45$ & $[-125.12, -75.83]$\\
Output tokens (k) & 0.188 & 0.201 & $+0.012$ & $[-0.003, 0.032]$\\
First-person images & 170.54 & 70.69 & $-99.84$ & $[-113.24, -86.80]$\\
Website screenshots & 7.75 & 8.44 & $+0.69$ & $[0.26, 1.13]$\\
Wall-clock time (s) & 279.93 & 159.47 & $-120.46$ & $[-160.35, -71.68]$\\
\end{appendixcostpanel}\\[3pt]
\begin{appendixcostpanel}
\multicolumn{5}{l}{\textbf{{Gemini}}}\\
Browser operations & 9.12 & 8.62 & $-0.49$ & $[-1.03, 0.04]$\\
Decision steps & 10.43 & 10.00 & $-0.43$ & $[-0.96, 0.09]$\\
Input tokens (k) & 502.60 & 310.90 & $-191.70$ & $[-236.83, -149.32]$\\
Output tokens (k) & 9.629 & 6.700 & $-2.929$ & $[-4.059, -1.879]$\\
First-person images & 233.67 & 88.81 & $-144.87$ & $[-173.96, -121.17]$\\
Website screenshots & 10.43 & 10.00 & $-0.43$ & $[-0.96, 0.09]$\\
Wall-clock time (s) & 229.10 & 161.41 & $-67.68$ & $[-88.93, -48.18]$\\
\end{appendixcostpanel}&
\begin{appendixcostpanel}
\multicolumn{5}{l}{\textbf{{qwen3.5-27b}}}\\
Browser operations & 9.21 & 8.41 & $-0.80$ & $[-1.77, 0.24]$\\
Decision steps & 10.15 & 9.44 & $-0.71$ & $[-1.67, 0.30]$\\
Input tokens (k) & 425.37 & 288.44 & $-136.93$ & $[-191.83, -80.67]$\\
Output tokens (k) & 0.156 & 0.155 & $-0.001$ & $[-0.014, 0.013]$\\
First-person images & 224.38 & 85.14 & $-139.24$ & $[-163.93, -116.74]$\\
Website screenshots & 10.15 & 9.44 & $-0.71$ & $[-1.67, 0.30]$\\
Wall-clock time (s) & 274.48 & 149.17 & $-125.31$ & $[-157.82, -92.62]$\\
\end{appendixcostpanel}\\[3pt]
\end{appendixlongtable}
\endgroup

Table~\ref{tab:successful-costs} checks whether the input savings persist when both runs succeed. Both comparisons use the recording-level analysis in Section~\ref{app:experiment-details}.

\begingroup\setlength{\tabcolsep}{0pt}
\begin{appendixlongtable}{p{.488\linewidth}p{.488\linewidth}}
\caption{{Mean resource use on tasks successful in both Offline and Online. The subset count $n$ is drawn from 200 pairs per model. Differences are Online minus Offline with recording-level bootstrap 95\% intervals.}}
\label{tab:successful-costs}\\
\endfirsthead
\appendixcontinued{2}
\endhead
\begin{appendixcostpanel}
\multicolumn{5}{l}{\textbf{{Terra, $n=163$}}}\\
Browser operations & 7.82 & 7.98 & $+0.16$ & $[-0.23, 0.55]$\\
Decision steps & 8.84 & 9.26 & $+0.42$ & $[0.00, 0.83]$\\
Input tokens (k) & 387.72 & 264.61 & $-123.11$ & $[-150.96, -96.05]$\\
Output tokens (k) & 0.415 & 0.432 & $+0.018$ & $[-0.029, 0.056]$\\
First-person images & 196.74 & 82.19 & $-114.55$ & $[-132.55, -98.13]$\\
Website screenshots & 8.84 & 9.26 & $+0.42$ & $[0.00, 0.83]$\\
Wall-clock time (s) & 100.41 & 75.34 & $-25.07$ & $[-33.22, -17.47]$\\
\end{appendixcostpanel}&
\begin{appendixcostpanel}
\multicolumn{5}{l}{\textbf{{DeepSeek, $n=105$}}}\\
Browser operations & 6.65 & 6.66 & $+0.01$ & $[-0.55, 0.54]$\\
Decision steps & 9.17 & 9.48 & $+0.30$ & $[-0.43, 1.04]$\\
Input tokens (k) & 258.18 & 198.61 & $-59.57$ & $[-93.92, -24.28]$\\
Output tokens (k) & 3.840 & 4.346 & $+0.506$ & $[-1.115, 2.566]$\\
First-person images & 190.76 & 80.16 & $-110.60$ & $[-133.41, -91.14]$\\
Website screenshots & 9.17 & 9.48 & $+0.30$ & $[-0.43, 1.04]$\\
Wall-clock time (s) & 57.06 & 55.05 & $-2.02$ & $[-17.01, 12.91]$\\
\end{appendixcostpanel}\\[3pt]
\begin{appendixcostpanel}
\multicolumn{5}{l}{\textbf{{Doubao, $n=132$}}}\\
Browser operations & 6.09 & 6.55 & $+0.46$ & $[-0.06, 0.98]$\\
Decision steps & 7.08 & 7.78 & $+0.70$ & $[0.17, 1.21]$\\
Input tokens (k) & 317.79 & 195.70 & $-122.08$ & $[-166.00, -82.05]$\\
Output tokens (k) & 1.339 & 1.172 & $-0.167$ & $[-0.493, 0.134]$\\
First-person images & 147.77 & 59.11 & $-88.66$ & $[-105.05, -73.53]$\\
Website screenshots & 7.08 & 7.78 & $+0.70$ & $[0.17, 1.21]$\\
Wall-clock time (s) & 193.07 & 101.79 & $-91.28$ & $[-119.59, -60.27]$\\
\end{appendixcostpanel}&
\begin{appendixcostpanel}
\multicolumn{5}{l}{\textbf{{Kimi, $n=58$}}}\\
Browser operations & 6.03 & 7.00 & $+0.97$ & $[0.33, 1.67]$\\
Decision steps & 7.05 & 8.10 & $+1.05$ & $[0.43, 1.72]$\\
Input tokens (k) & 281.61 & 200.39 & $-81.23$ & $[-111.72, -49.95]$\\
Output tokens (k) & 0.181 & 0.183 & $+0.002$ & $[-0.012, 0.016]$\\
First-person images & 145.95 & 62.84 & $-83.10$ & $[-104.88, -63.48]$\\
Website screenshots & 7.05 & 8.10 & $+1.05$ & $[0.43, 1.72]$\\
Wall-clock time (s) & 260.76 & 132.74 & $-128.02$ & $[-173.96, -88.47]$\\
\end{appendixcostpanel}\\[3pt]
\begin{appendixcostpanel}
\multicolumn{5}{l}{\textbf{{Gemini, $n=133$}}}\\
Browser operations & 7.79 & 7.44 & $-0.35$ & $[-0.92, 0.23]$\\
Decision steps & 9.15 & 8.83 & $-0.32$ & $[-0.90, 0.28]$\\
Input tokens (k) & 422.13 & 248.45 & $-173.67$ & $[-230.68, -126.17]$\\
Output tokens (k) & 7.817 & 4.911 & $-2.906$ & $[-4.281, -1.717]$\\
First-person images & 210.24 & 77.50 & $-132.74$ & $[-173.44, -104.83]$\\
Website screenshots & 9.15 & 8.83 & $-0.32$ & $[-0.90, 0.28]$\\
Wall-clock time (s) & 194.12 & 128.71 & $-65.41$ & $[-86.33, -44.37]$\\
\end{appendixcostpanel}&
\begin{appendixcostpanel}
\multicolumn{5}{l}{\textbf{{qwen3.5-27b, $n=56$}}}\\
Browser operations & 8.04 & 7.82 & $-0.21$ & $[-1.73, 1.32]$\\
Decision steps & 9.04 & 8.89 & $-0.14$ & $[-1.64, 1.41]$\\
Input tokens (k) & 348.31 & 249.84 & $-98.48$ & $[-179.81, -19.89]$\\
Output tokens (k) & 0.141 & 0.143 & $+0.002$ & $[-0.016, 0.020]$\\
First-person images & 177.73 & 74.07 & $-103.66$ & $[-132.91, -74.89]$\\
Website screenshots & 9.04 & 8.89 & $-0.14$ & $[-1.64, 1.41]$\\
Wall-clock time (s) & 241.93 & 139.28 & $-102.66$ & $[-168.46, -49.59]$\\
\end{appendixcostpanel}\\[3pt]
\end{appendixlongtable}
\endgroup

The same task subsets are used for the Computer-Use input reference. Its mean input tokens, in thousands, are 127.15 for Terra, 106.84 for Doubao, 157.16 for Gemini, 151.39 for DeepSeek, 75.87 for Kimi, and 132.67 for Qwen. Subset membership is determined by Offline and Online success.

On tasks successful in both settings, Online still uses fewer input tokens and first-person images for every model. Online delivers the recording gradually and compresses older images, while Offline repeatedly supplies the full recording. Lower visual input does not imply fewer browser operations: Kimi makes more, and the other models show no clear difference. Wall-clock duration is lower for five models, with no clear difference for DeepSeek.

\section{Trajectory Analysis}
\label{app:trajectory-analysis}

\subsection{Error Categories}
Figure~6 of the main paper groups failures by the three challenges in the Introduction. Table~\ref{tab:failure} defines the five underlying error categories. To assign a category, we align the delivered observations and requests with browser actions, feedback, and website state. We label the earliest error that contributes to failure and remains uncorrected.

\begin{appendixcompacttable}
\centering
\caption{Failure categories and their observable distinctions in trajectory analysis.}
\label{tab:failure}
\small
\renewcommand{\arraystretch}{1.12}
\begin{tabular*}{\linewidth}{@{\extracolsep{\fill}}lp{.77\linewidth}@{}}
\toprule
Category & Observable distinction\\
\midrule
Perception & {Behavior adopts a visual property contradicted by already delivered frames.}\\
Intent & {The agent commits to an unresolved referent or explicitly follows an obsolete requirement.}\\
Linking & {The selected webpage entity conflicts with the active request, or a compatible candidate is explicitly rejected.}\\
State Control & {An invalid action or unrecovered execution error occurs, or an evidenced target is not realized through state edits.}\\
Verification & {An observable completion error causes failure, such as destroying a correct state without recovery.}\\
\bottomrule
\end{tabular*}
\end{appendixcompacttable}

{For presentation, Perception and Linking are associated primarily with Context Grounding, Intent with Intent Continuity, and State Control and Verification with Execution Adaptation.}

\subsection{Attribution Procedure}
The final-state checker determines which runs fail. The read-only gpt-6-luna Judge then assigns an error category using the delivered requests and frames, public agent responses, browser observations, action feedback, and website state. The Judge receives the scoring goal and reference answer separately from the evidence available to the agent at each step. It identifies the decisive step and its supporting evidence, or leaves the category unassigned when the evidence is insufficient.

We retry failed Judge calls before human review. A reviewer uses task and trajectory summaries to assign a label when the Judge result is uncertain or unavailable. Final-state scores are unchanged.

\Needspace{28\baselineskip}
\subsection{Failure Counts}
Table~\ref{tab:failure-counts} gives the breakdown for the 1,263 runs with an error label out of 1,505 failures. Figure~6 of the main paper shows the 604 labeled failures out of 807 for Terra, DeepSeek, Kimi, and qwen3.5-27b across the three settings.

\begin{table*}
\centering\small
\caption{Labeled failures by model and setting. Human-labeled counts are included in the five error columns.}
\label{tab:failure-counts}
\renewcommand{\arraystretch}{1.12}
\setlength{\tabcolsep}{3pt}
\begin{tabular*}{\textwidth}{@{\extracolsep{\fill}}llrrrrrr@{}}
\toprule
 & & \multicolumn{5}{c}{Assigned error} & \\
\cmidrule(lr){3-7}
Model & Setting & Perception & Intent & Linking & State Control & Verification & Human-labeled\\
\midrule
Terra & Computer-Use & 0 & 0 & 2 & 2 & 0 & 0\\
Terra & Offline & 2 & 14 & 4 & 1 & 0 & 1\\
Terra & Online & 2 & 9 & 1 & 2 & 0 & 1\\
\addlinespace
Gemini & Online & 2 & 11 & 8 & 6 & 1 & 1\\
\addlinespace
Doubao & Online & 2 & 16 & 11 & 4 & 0 & 2\\
\addlinespace
DeepSeek & Computer-Use & 0 & 0 & 5 & 4 & 0 & 0\\
DeepSeek & Offline & 14 & 20 & 8 & 7 & 0 & 0\\
DeepSeek & Online & 9 & 23 & 16 & 6 & 2 & 5\\
\addlinespace
Kimi & Computer-Use & 0 & 0 & 30 & 0 & 0 & 0\\
Kimi & Offline & 8 & 32 & 55 & 2 & 0 & 4\\
Kimi & Online & 12 & 31 & 48 & 5 & 1 & 4\\
\addlinespace
Qwen & Computer-Use & 0 & 0 & 37 & 1 & 1 & 1\\
Qwen & Offline & 9 & 37 & 44 & 3 & 2 & 1\\
Qwen & Online & 10 & 28 & 47 & 8 & 0 & 5\\
\addlinespace
UI-TARS & Computer-Use & 0 & 0 & 0 & 199 & 0 & 0\\
UI-TARS & Offline & 0 & 0 & 0 & 200 & 0 & 0\\
UI-TARS & Online & 0 & 0 & 0 & 199 & 0 & 0\\
\midrule
Total & All available & 70 & 221 & 316 & 649 & 7 & 25\\
\bottomrule
\end{tabular*}
\end{table*}

\paragraph{Revisions after an update.}
Table~\ref{tab:update-state} shows that in 149 of the 1,019 Online Ego Use runs that receive a Graft or Refresh update, the website satisfies the initial goal but not the revised goal at delivery. Of these 149 runs, 107 reach the revised goal and 42 fail. Among the 26 labeled failures, Intent accounts for nine, and Linking and State Control account for eight each.

\Needspace{36\baselineskip}
\subsection{Case Studies}
\label{app:case-evidence}

Table~\ref{tab:case-trajectories} uses two comparisons to show what the final website state leaves out. On the candy task, four agents make the same selection with different evidence about the user's intended item. On the repainting task, two runs of Terra find the same suitable tutorial but end with different playlists.

\paragraph{Referent resolution.}
In the audit in Section~\ref{app:clarification-behavior}, 69 of the 78 successful Online runs make their final selection while two or more interpretations still require different website choices. Clarification matters when the scene narrows the possibilities but does not distinguish the remaining options.

The user in the candy task asks for the candy \emph{right next to} a blue Airheads bag. Trolli is on the left and Russell Stover on the right, so proximity alone does not identify one of them. Terra asks about the two options at step 1 and receives a reply specifying the left side before step 2. It adds Trolli at step 5. The other agents add Trolli without asking. The checker accepts the same cart in all four runs, but only Terra receives information that rules out the other candy.

In \texttt{CR-21}, the user requests a spare preparation tool and the next two ingredients added to a salad. Both DeepSeek and Gemini choose suitable ingredients from the later frames. DeepSeek adds a cutting board without asking which tool the user means. Gemini asks at the first step, receives a description of the scale behind the cutting board, and adds a scale. The ingredient updates guide both runs, but only Gemini resolves the separate reference to the tool.

\paragraph{Website state revision.}
Among the ten Terra pairs in Section~\ref{app:online-input-conditions} where Concurrent succeeds and Wake-up fails, five Wake-up runs find a complete new answer but retain an extra entry selected before the follow-up. The agent acts on the revised request without fully revising the website state.

On the repainting task, both Terra runs start with an empty playlist. The initial request is for a repainting tutorial. The follow-up asks for one that also covers surface preparation. Offline, Terra receives both requests before acting and saves a suitable guide at step 8. Online, Terra saves a guide at step 4, receives the follow-up at step 6, and saves the same suitable guide as the Offline run at step 11. It stops at step 12 of the 20-step budget with both guides still saved. The checker accepts a playlist containing one suitable guide. Removing the first guide would make the Online playlist identical to the accepted Offline playlist.

Revision can also require keeping part of the earlier state. In \texttt{CRG-14}, Gemini has already added a tea strainer and a citrus squeezer when the user asks to replace the squeezer with bottled juice. Gemini removes the squeezer and adds the juice, leaving the tea strainer in the cart as requested.

\begin{table}
\small
\renewcommand{\arraystretch}{1.12}
\setlength{\tabcolsep}{5pt}
\caption{Key decisions and final states for the candy and repainting tasks. Arrows show the order of events. Panel (b) compares Terra under Offline and Online Ego Use with the same final-state scoring criterion.}
\label{tab:case-trajectories}
\newcommand{\casearrow}{\enspace\ensuremath{\rightarrow}\enspace}

\noindent\textbf{(a) Candy selection}\hfill\texttt{candidate-mart-clarify-position-02}\par\smallskip
\begin{tabularx}{\linewidth}{@{}>{\raggedright\arraybackslash}p{.17\linewidth}>{\raggedright\arraybackslash}X >{\raggedright\arraybackslash}p{.23\linewidth}l@{}}
\toprule
Model & Clarification before selection & Final cart & Outcome\\
\midrule
Terra & Asks at step 1\casearrow Reply specifies the left side & Trolli & Pass\\
Gemini, Doubao, DeepSeek & No clarification request & Trolli & Pass\\
\bottomrule
\end{tabularx}

\medskip
\noindent\textbf{(b) Repainting tutorial}\hfill\texttt{candidate-media-graft-c-56}\par\smallskip
\begin{tabularx}{\linewidth}{@{}>{\raggedright\arraybackslash}p{.17\linewidth}>{\raggedright\arraybackslash}X >{\raggedright\arraybackslash}p{.23\linewidth}l@{}}
\toprule
Setting & Sequence of events & Final playlist & Outcome\\
\midrule
Offline Ego Use & Both requests available\casearrow Saves guide at step 8 & Updated tutorial only & Pass\\
Online Ego Use & Saves at step 4\casearrow Follow-up at step 6\casearrow Saves updated guide at step 11 & Earlier and updated tutorials & Fail\\
\bottomrule
\end{tabularx}
\end{table}

\end{document}